\documentclass[lettersize,journal]{IEEEtran}
\usepackage{amsmath,amsfonts}
\usepackage{algorithmic}
\usepackage{algorithm}
\usepackage{array}
\usepackage[caption=false,font=normalsize,labelfont=sf,textfont=sf]{subfig}
\usepackage{textcomp}
\usepackage{stfloats}
\usepackage{url}
\usepackage{verbatim}
\usepackage{graphicx}
\usepackage{cite}
\usepackage{color} 
\usepackage{caption}
\usepackage{multirow}
\usepackage{booktabs}
\usepackage[table,xcdraw]{xcolor} 
\usepackage[colorlinks,bookmarksopen,bookmarksnumbered,citecolor=blue, linkcolor=red, urlcolor=blue]{hyperref}
\begin{document}

\title{TalkingEyes: Pluralistic Speech-Driven 3D Eye Gaze Animation}

\author{Yixiang Zhuang, Chunshan Ma, Yao Cheng, Xuan Cheng, Jing Liao, Juncong Lin 
\thanks{Yixiang Zhuang, Xuan Cheng and Juncong Lin are with the School of Informatics, Xiamen University, China. E-mail: yixiangzhuang@stu.xmu.edu.cn, chengxuan@xmu.edu.cn, jclin@xmu.edu.cn.}
\thanks{Chunshan Ma and Yao Cheng are with China Mobile (Hangzhou) Information Technology Co., Ltd., China. E-mail: machunshan@cmhi.chinamobile.com, chengyao@cmhi.chinamobile.com.}
\thanks{Jing Liao is with the Department of Computer Science, City University of Hong Kong, Hong Kong. E-mail: jingliao@cityu.edu.hk.}
\thanks{Corresponding author: Xuan Cheng.}
}

\maketitle

\markboth{Journal of \LaTeX\ Class Files,~Vol.~14, No.~8, August~2021}%
{Shell \MakeLowercase{\textit{et al.}}: A Sample Article Using IEEEtran.cls for IEEE Journals}


\begin{abstract}
Although significant progress has been made in the field of speech-driven 3D facial animation recently, the speech-driven animation of an indispensable facial component, eye gaze, has been overlooked by recent research. This is primarily due to the weak correlation between speech and eye gaze, as well as the scarcity of audio-gaze data, making it very challenging to generate 3D eye gaze motion from speech alone. In this paper, we propose a novel data-driven method which can generate diverse 3D eye gaze motions in harmony with the speech. To achieve this, we firstly construct an audio-gaze dataset that contains about 14 hours of audio-mesh sequences featuring high-quality eye gaze motion, head motion and facial motion simultaneously. The motion data is acquired by performing lightweight eye gaze fitting and face reconstruction on videos from existing audio-visual datasets. We then tailor a novel speech-to-motion translation framework in which the head motions and eye gaze motions are jointly generated from speech but are modeled in two separate latent spaces. This design stems from the physiological knowledge that the rotation range of eyeballs is less than that of head. Through mapping the speech embedding into the two latent spaces, the difficulty in modeling the weak correlation between speech and non-verbal motion is thus attenuated. Finally, our TalkingEyes, integrated with a speech-driven 3D facial motion generator, can synthesize eye gaze motion, eye blinks, head motion and facial motion collectively from speech. Qualitative and quantitative evaluations, along with a perceptual user study, demonstrate the superiority of the proposed method in generating diverse and natural 3D eye gaze motions from speech. The project page of this paper is: \href{https://lkjkjoiuiu.github.io/TalkingEyes_Home/}{https://lkjkjoiuiu.github.io/TalkingEyes\_Home/}.

\end{abstract}


\section{Introduction}
The computer graphics and vision community has witnessed a significant progress in the field of speech-driven 3D facial animation, largely due to the recent boom in generative models \cite{vae2014, VQVAE2017,  GANs2014, StyleGAN2019, Attention2017, DDPM2020, StableDiffusion, DiffSFSR2024} and audio-visual datasets \cite{chung2018voxceleb2, wang2020mead, zhang2021flow, VOCA2019, BIWI2010}. 
Recent speech-driven 3D facial animation research is dedicated to generating mouth movements \cite{VOCA2019, MeshTalk2021, FaceFormer2022, CodeTalker2023, FaceDiffuser2023, DNPMICME2024, Learn2Talk2025}, head gestures \cite{3DTalkingFaceHead2023, DiffPoseTalkTOG2024, Learn2Talk2025} and emotional facial expressions \cite{EmoTalk2023, EMOTE2023, Media2FaceSIG2024}, solely from speech signals. A considerable focus is on achieving synchronization between the animated mouth movements and the corresponding speech. However, the speech-driven animation of an indispensable facial component, eye gaze, has been overlooked by the recent research. Eyes, our ``windows to the soul", serve a crucial role in indicating an individual's emotions, thoughts, feelings and intentions, thus enabling us to ``see someone's soul" through observing their eyes. In the absence of animated eye gaze, the synthesized 3D talking faces \cite{VOCA2019, MeshTalk2021, FaceFormer2022, CodeTalker2023, FaceDiffuser2023, 3DTalkingFaceHead2023, DiffPoseTalkTOG2024, DNPMICME2024, Learn2Talk2025, EmoTalk2023, EMOTE2023, Media2FaceSIG2024} tend to look straight ahead with a static eye pose, resulting in an unnatural and wooden looking facial behavior. Although the recent scene-driven methods \cite{SaliencyGazeTVCG2024, S3TOG2024} achieve impressive eye gaze animation for virtual characters, they require other multi-modal cues besides speech as input, such as scene image, scene context, director script and others.

In this work, we focus on task of generating the expressive 3D eye gaze motion from speech alone, which is in practice challenging for two reasons. 1) \emph{\textbf{Weak Correlation}}. Eye gaze is one of many non-verbal cues used in communication. It is possible for a speaker to direct his gaze elsewhere while still effectively communicating his message via other verbal means. Mouth strongly correlates with speech, while eye gaze is weakly correlated. As a result, there exists a cross-modal one-to-many mapping between speech signal and eye gaze motion, which is in practice hard to be captured. 2) \emph{\textbf{Data Scarcity}}. Existing audio-mesh datasets including VOCASET \cite{VOCA2019}, BIWI \cite{BIWI2010}, Speech4Mesh \cite{he2023speech4mesh}, Audio2Mesh \cite{yang2024probabilistic} and DiffPoseTalk \cite{DiffPoseTalkTOG2024} focus on collecting facial motions (mainly mouth movements), but neglect the eye gaze capturing. A 3D dataset consisted of eye gaze motion, head motion and facial motion and their corresponding speech recordings is really scarce.

When we revisit the early research on computer animation, we find that pioneers in this field had proposed a few potential solutions \cite{HeadEyeAnimation2007, VirtualCharacterSCA2013, ConversationalAgents2012, LiveSpeehEyeTVCG2012, EyeMotion2019}, roughly divided into rule-based \cite{ HeadEyeAnimation2007, VirtualCharacterSCA2013} and learning-based methods \cite{ConversationalAgents2012, LiveSpeehEyeTVCG2012, EyeMotion2019}. 
However, both types of methods can only establish a one-to-one mapping between speech signal and eye gaze motion, resulting in deterministic and over-smoothed motion. Additionally, their motion capture systems used to collect eye gaze data in the lab, are costly and time-consuming for constructing large-scale in-the-wild dataset. Ultimately, these methods are mostly customized for specific avatars, making them difficult to integrate with generic 3D face models like FLAME \cite{FLAME2017}.

\begin{figure*}
    \centering
    \includegraphics[width=1.0\textwidth]{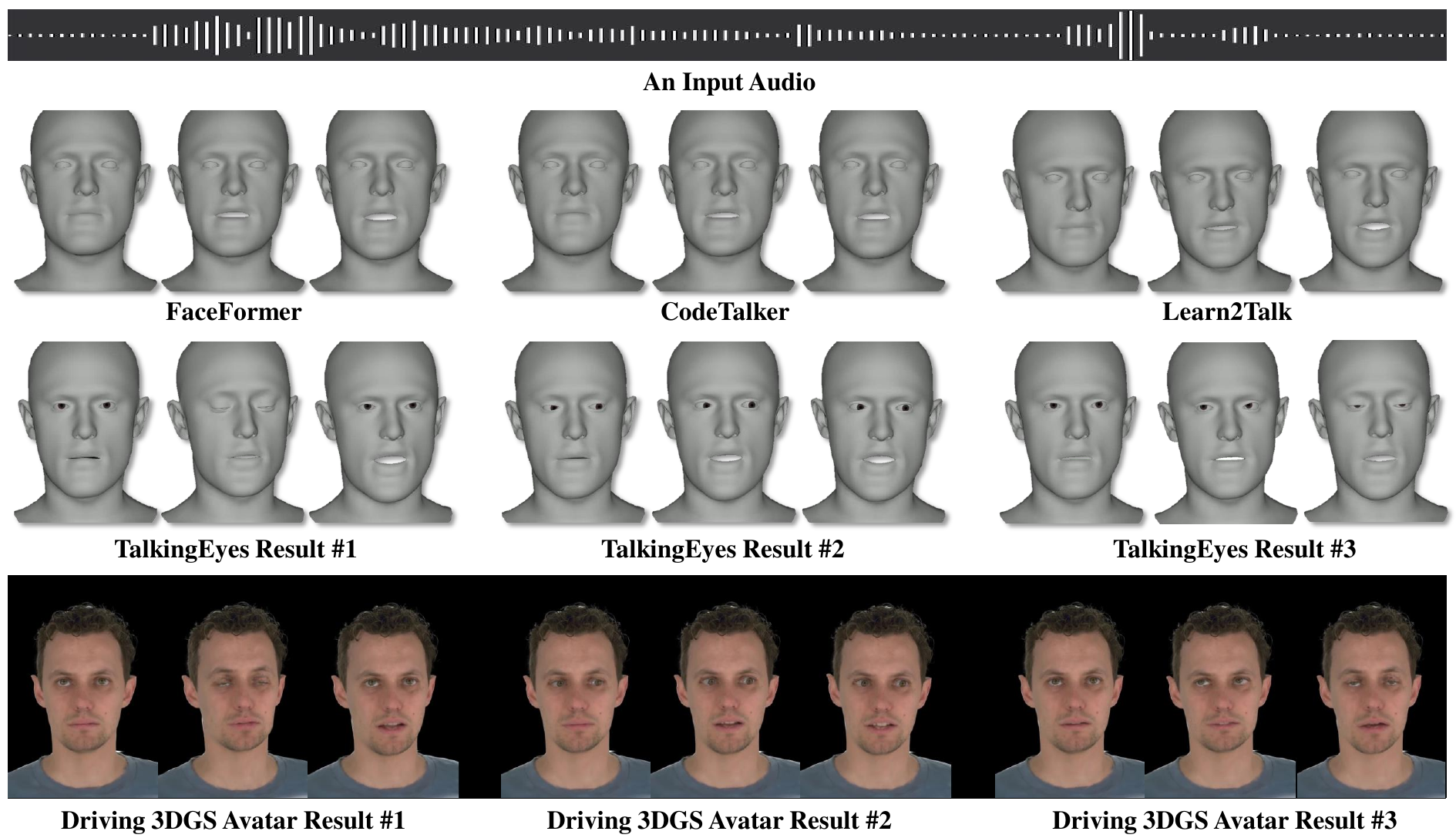}
    \caption{Given a speech audio, our TalkingEyes can synthesize pluralistic eye gaze motions, eye blinks and head motions and facial motions, and further be used to drive a 3D Gaussian Splatting (3DGS) \cite{3DGSTOG2023} based head avatar \cite{GaussianAvatarsCVPR2024}.}
    \label{fig:teaser}
\end{figure*}

To address the above challenges in task and the limitations in existing methods, we propose \emph{\textbf{TalkingEyes}}, a data-driven method which can generate pluralistic 3D eye gaze motions in tune with the speech. To this end, we firstly construct 3D \textbf{\emph{T}}al\textbf{\emph{K}}ing \textbf{\emph{E}}yes \textbf{\emph{D}}ataset (\emph{\textbf{TKED}}) by conducting 3D eye gaze fitting and 3D face reconstruction on the videos from audio-visual datasets to obtain pseudo ground truth. A lightweight 3D eye gaze fitting method named \emph{LightGazeFit} is tailored to estimate the eyeball rotation from the low-quality Internet videos. TKED contains 14 hours of high-quality audio-mesh sequences featuring eye gaze, head and facial motions, thus  facilitating the generation of 3D talking eyes, 3D talking head and 3D talking face simultaneously.

Secondly, we propose to model the eye gaze motion and head motion across two latent spaces, and subsequently predict them from speech via a temporal autoregressive model. We employ a variational autoencoder (VAE) to learn continuous latent space for head motion and a vector-quantized variational autoencoder (VQVAE) to learn discrete latent space for eye gaze motion. This distinction is rooted in the different degrees of motion diversity, caused by the different rotation ranges of eyeballs and head. The head motion is firstly generated from speech and then used as a condition in the autoregressive generation of eye gaze motion, with respect to the synchronous nature of head and eye movements \cite{Ophthal1972, Warabi1977, LiveSpeehEyeTVCG2012}. Qualitative and quantitative evaluations, along with a perceptual user study, demonstrate the superiority of the proposed TalkingEyes in generating diverse and natural 3D eye gaze motions from speech.

To create holistic animation, the eye blinks are synthesized based on a statistical analysis on TKED, and the mouth and upper-face movements are generated by a speech-driven facial animation method \cite{Learn2Talk2025}.

The contributions of this paper are summarized as follows:
\begin{itemize}
    \item TalkingEyes is the first holistic 3D talking avatar that can automatically and collectively generate eye gaze motion, eye blinks, head motion and facial motion from speech alone, as shown in Fig. \ref{fig:teaser}.

    \item TKED is the first audio-mesh dataset that contains three types of resources for speech-driven animation research: 3D talking eyes, 3D talking head and 3D talking face.
    
    \item VAE and VQVAE are proposed to model different degrees of diversity in head and eye gaze motions. This facilitates the generation of non-deterministic, diverse and natural eye gaze motions, in contrast with the deterministic generation in existing methods.

    \item LightGazeFit is proposed to process large-scale low-quality internet videos, without the requirement of accurate segmentation and user-specific calibration as existing 3D eye gaze fitting methods.

\end{itemize}


\section{Related Work}
The 3D eye gaze animation can be driven by audio, image and video, and is also close related to 3D facial animation. We introduce representative works in these fields, and recommend a comprehensive review \cite{EyeGazeReview2015} of inter-disciplinary research on eye gaze.

\subsection{Speech-Driven 3D Eye Gaze Animation}
It's a challenging task to generate 3D eye gaze animation from speech, and most existing methods were proposed a decade ago. The methods can be roughly categorized into rule-based \cite{ HeadEyeAnimation2007, VirtualCharacterSCA2013} and learning-based \cite{ConversationalAgents2012, LiveSpeehEyeTVCG2012, EyeMotion2019}. 

Rule-based methods construct the mapping between conversation states and eye gaze patterns by defining animation rules. Masuko and Hoshino \cite{HeadEyeAnimation2007} designed a set of empirical rules to map the conversation states (e.g. utterance, listening and waiting) to the amount of gaze and gaze duration. Marsella et al. \cite{VirtualCharacterSCA2013} proposed a rule system to sequentially convert the audio and text features to communicative functions, behavior classes, specific behaviors and finally animation. 

In contrast, learning-based methods capture the relationship between speech signal and eye movements by training on data. Mariooryad and Busso \cite{ConversationalAgents2012} employed three Dynamic Bayesian Networks to model the coupling between speech, eyebrow and head motion. Le et al. \cite{LiveSpeehEyeTVCG2012} utilized non-linear dynamical canonical correlation analysis to synthesize gaze from head motion and speech features. Jin et al. \cite{EyeMotion2019} proposed a LSTM-based machine learning model to predict the direction-of-focus of all the interlocutors in a three-party conversation. All the above learning-based methods can generate head-eye animation from speech, but they only establish a one-to-one mapping between speech feature and eye gaze motion, without the modeling of uncertainty in nonverbal communication.

To reduce uncertainty, some recent scene-driven (not solely speech-driven) methods \cite{SaliencyGazeTVCG2024, S3TOG2024, migCanalesJJ23} forced the animated eye gaze to focus on the interlocutors or on the salient points within the scene. Our method only takes the speech as input and achieves the goal of generating plausible eye gaze motion in a speech-centric setting.

\subsection{Eye Gaze Fitting in Video}
Compared with speech, using video to generate eye gaze motion is more straightforward and convenient. The researchers in computer graphics have proposed several 3D face trackers for video. Extending 3D face reconstruction methods \cite{Deep3DFace2019, DECA2021, DCT2024, FaceRefiner2024, DNPMICME2024}, these trackers jointly fit parameters for both face and eye gaze states to image cues, enabling them to reconstruct a 3D face model along with 3D eye gaze for each frame. Wang et al. \cite{wang2016realtime} utilized a random forest classifier to extract the iris and pupil pixels in the eye region, and then used them to sequentially infer the most likely state of the 3D eye gaze at each frame in the MAP framework. Wang et al. \cite{wang2019realtime} improved the work \cite{wang2016realtime} by using CNN rather than random forest classifier for the iris and pupil pixels classification. Wen et al. \cite{wen2017real} introduced a method for estimating eyeball motions for RGBD inputs by minimizing the differences between a rendered eyeball and a recorded image. 

The above 3D eye gaze fitting methods require accurate segmentation of iris and pupil pixels and user-specific calibration, which is difficult to be achieved in low resolution videos.

\subsection{Gaze Estimation from Image}
The computer vision community starts very early to estimate gaze from image. The estimated gaze is usually in the form of a unit gaze direction vector in 3D space or a point of gaze (PoG) on 2D plane. Early gaze estimation methods are mainly template-matching-based, appearance-based and feature-based, while recent methods are mostly based on deep learning \cite{GazeEstimationSurvey2024}. Zhang et al. \cite{zhang2015appearance} introduced the first CNN-based method for gaze estimation, which jointly utilizes eye images and head poses to predict gaze directions. Accounting for the potential asymmetry between human eyes, Cheng et al. \cite{cheng2020gaze} proposed to firstly predict the 3D gaze direction for each eye individually and then employ a reliability evaluation network to adaptively adjust the weighting strategy throughout the optimization. Park et al.\cite{park2018deep} introduced a novel graphical representation of 3D gaze direction, which serves as intermediate supervision to improve the estimation accuracy. Owing to individual differences, some methods \cite{park2019few, yu2019improving} employed a few-shot approach to achieve user-specific gaze adaptation.
To address the challenge of occluded eyes, Kellnhofer et al. \cite{kellnhofer2019gaze360} utilized visible head features to predict gaze direction, even when the eyes are completely obscured, and indicated the prediction’s limited accuracy by outputting a higher uncertainty value accordingly. 

These deep-learning-based gaze estimation methods focus on eyes, without considering the relationship between gaze estimation and 3D face reconstruction.

\subsection{Speech-Driven 3D Facial Animation}
Methods in this field are roughly categorized into linguistics-based \cite{JALI2016, AnimatedSpeech2001, DynamicUnits2012, LipSyncGames2013} and learning-based \cite{VOCA2019, MeshTalk2021, FaceFormer2022, CodeTalker2023, FaceDiffuser2023, DiffPoseTalkTOG2024, Learn2Talk2025, EmoTalk2023, EMOTE2023, Media2FaceSIG2024}. Linguistics-based methods typically establish a set of mapping rules between phonemes and visemes, thus providing explicit control over the animation. JALI \cite{JALI2016} is a representative recent linguistics-based method, which utilizes two anatomical actions (jaw and lip) to animate a 3D facial rig. As a lot of manual effort is required for tuning animation parameters, a wide variety of data-driven methods have been proposed as an alternative. The generative models are utilized in these methods to directly learn the mapping from speech features to 3D facial motion, which include CNN \cite{VOCA2019, MeshTalk2021}, Transformer \cite{FaceFormer2022, CodeTalker2023, EmoTalk2023, EMOTE2023, Learn2Talk2025} and Diffusion Model \cite{FaceDiffuser2023, DiffPoseTalkTOG2024, Media2FaceSIG2024}. VOCASET \cite{VOCA2019} and BIWI \cite{BIWI2010} are two widely used 3D audio-visual datasets to train the audio-to-mesh regression networks in the learning-based methods.

Considering that Learn2Talk \cite{Learn2Talk2025} has achieved impressive lip-sync facial animation, we choose it as the 3D facial motion generator in our proposed method.

\begin{table*}[tp]
\centering
\fontsize{8.0}{8.0}\selectfont
\begin{tabular}{ccccccccccc}
	\midrule
	\multirow{3}{*}{Dataset}
        &\multicolumn{3}{c}{Motion Types}
        &\multirow{3}{*}{Year}
        &\multirow{3}{*}{\shortstack{Number\\of Sequences}}
        &\multirow{3}{*}{\shortstack{Number\\of Subjects}}
        &\multirow{3}{*}{Hours}
        &\multirow{3}{*}{\shortstack{Data Acquisition\\Mode}}
        &\multirow{3}{*}{Video Sources}\cr
        
	\cmidrule(l){2-4} 
	& Face& Head & Eyes\cr
	\midrule
	VOCASET\cite{VOCA2019}&\checkmark&-&-&2019&480&12&0.48&3D Scanning&/\cr
	BIWI\cite{BIWI2010}&\checkmark&-&-&2010&1109&14&1.43&3D Scanning&/\cr
	Speech4Mesh\cite{he2023speech4mesh}&\checkmark&-&-&2023&2000&/&/&Generated from video&MEAD\cite{wang2020mead}, VoxCeleb2\cite{chung2018voxceleb2}\cr
	Audio2Mesh\cite{yang2024probabilistic}&\checkmark&-&-&2024&
    \textgreater 1M&6112&/&Generated from video&VoxCeleb2\cite{chung2018voxceleb2}\cr
	DiffPoseTalk\cite{DiffPoseTalkTOG2024}&\checkmark&\checkmark&-&2024&1052&588&26.5&Generated from video&TFHP\cite{DiffPoseTalkTOG2024}, HDTF\cite{zhang2021flow}\cr
    
    \midrule
    \textcolor[rgb]{0.05,0.5,0.7}{Our TKED}&\textcolor[rgb]{0.05,0.5,0.7}\checkmark&\textcolor[rgb]{0.05,0.5,0.7}\checkmark&\textcolor[rgb]{0.05,0.5,0.7}\checkmark&\textcolor[rgb]{0.05,0.5,0.7}{2025}&\textcolor[rgb]{0.05,0.5,0.7}{5982}&\textcolor[rgb]{0.05,0.5,0.7}{669}&\textcolor[rgb]{0.05,0.5,0.7}{14.0}&\textcolor[rgb]{0.05,0.5,0.7}{Generated from video}&\textcolor[rgb]{0.05,0.5,0.7}{VoxCeleb2\cite{chung2018voxceleb2}, HDTF \cite{zhang2021flow}}\cr
    \midrule
\end{tabular}
\caption{Comparison of different audio-mesh datasets.}
\label{tab:dataset_compare}
\end{table*}

\begin{figure*}
    \centering
    \includegraphics[width=1.0\textwidth]{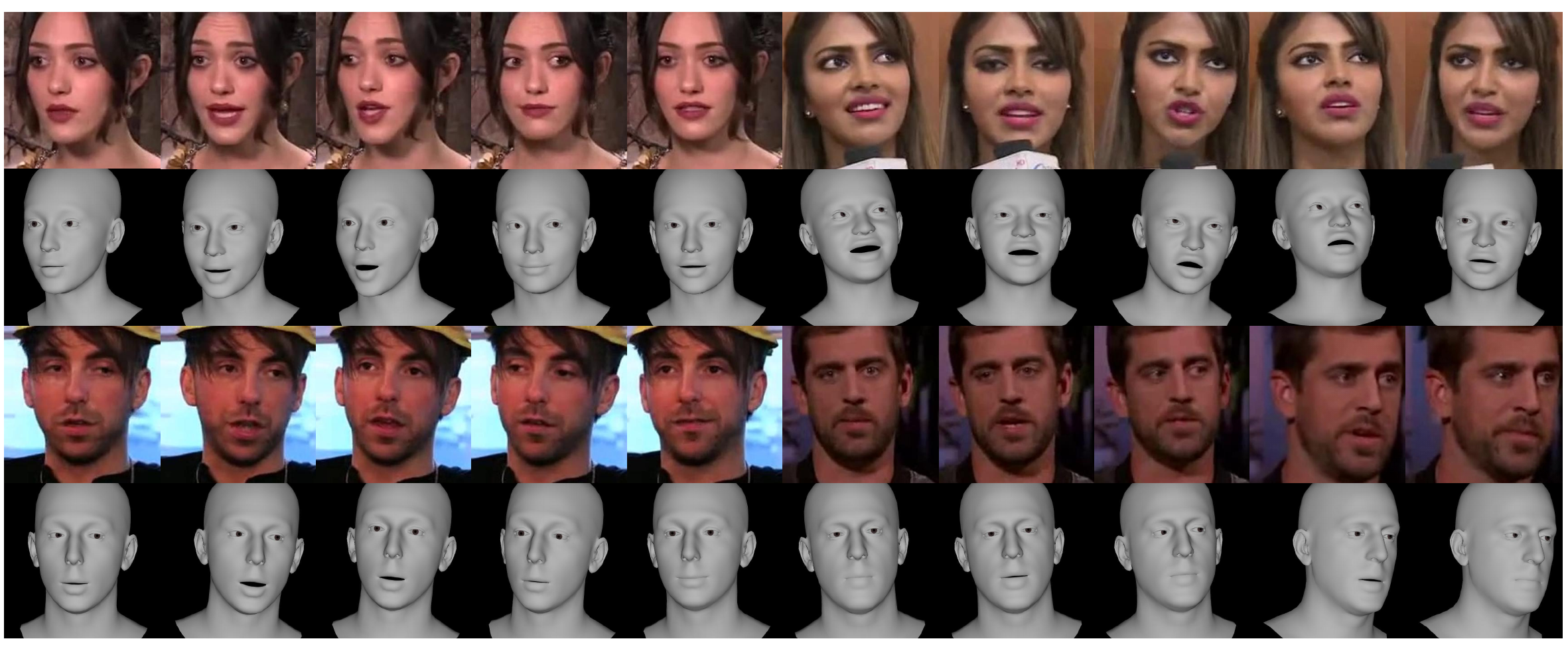}
    \caption{The videos and their reconstructed 3D mesh sequences in TKED.}
    \label{fig:dataset_presentation}
\end{figure*}

\section{Dataset: TKED}
Existing audio-mesh datasets have focused on collecting 3D facial motions isolated from eye gaze. To develop a learning-based 3D eye gaze motion generation method, it is essential to firstly construct an audio-gaze dataset with a sufficient amount of data and diverse subjects. We provide detailed description of our proposed dataset TKED in Sect. \ref{sect:dataset_description}, the construction pipeline of TKED in Sect. \ref{sect:dataset_pipeline} and the 3D eye gaze fitting method in Sect. \ref{sect:LightGazeFit}. 

\subsection{Dataset Description}
\label{sect:dataset_description}
The way of utilizing motion capture systems or eye tracking devices to capture eye gaze and microphone to record audio in a laboratory setting is limited in its ability to collect large-scale and diverse data. Considering that 2D audio-visual datasets are more accessible and offer broader coverage, it's possible to construct the audio-gaze dataset by reconstructing both 3D face mesh and 3D eye gaze from video. Motivated by this, we propose TKED which is obtained by performing 3D face reconstruction and 3D eye gaze fitting on the VoxCeleb2 \cite{chung2018voxceleb2} and HDTF \cite{zhang2021flow} datasets. 

VoxCeleb2 is a large-scale dataset collected from YouTube, comprising short interview videos cropped and resized to a $224\times224$ resolution. The speakers in the dataset span a wide range of different ethnicities, accents, professions and ages. A distinct aspect of the VoxCeleb2 videos is the frequent movement of the speakers' eyeballs, which contributes to a diverse range of eyeball rotation patterns. To mitigate the potential bias in TKED towards the specific eyeball movement style present in the VoxCeleb2 videos, we have added a small number of videos from HDTF as the complement. HDTF is a high-quality collection of talking face videos of 720P or 1080P, carefully gathered from YouTube. In contrast to VoxCeleb2, the videos from HDTF exhibit relatively stationary eyeballs.


TKED consists of 5,982 videos featuring 669 subjects, amounting to approximately 14 hours of footage in total. Out of the 5,982 videos, 5,912 are from VoxCeleb2, with a total duration of about 12 hours, while the remaining 70 videos are from HDTF, totaling about 2 hours. To ensure consistency, the videos from HDTF have been resampled to match the 25 fps in VoxCeleb2. We have conducted 3D face reconstruction and 3D eye gaze fitting on each frame of these videos to obtain the pseudo ground truth FLAME parameters \cite{FLAME2017}, which include two eyeballs' rotation, head rotation and facial motion parameters. Fig. \ref{fig:dataset_presentation} shows four sequences of the reconstructed 3D FLAME model. Along with the audio stored in the original videos, TKED is first dataset in the research community to simultaneously contain the three types of resources: 3D talking eyes, 3D talking head and 3D talking faces. The comprehensive comparison between TKED and existing audio-mesh datasets \cite{VOCA2019, BIWI2010, he2023speech4mesh, yang2024probabilistic, DiffPoseTalkTOG2024} is presented in Tab. \ref{tab:dataset_compare}. The training, validation and testing data in TKED are split in an 8:1:1 ratio based on the number of video clips.

\begin{figure*}
    \centering
    \includegraphics[width=1.0\textwidth]{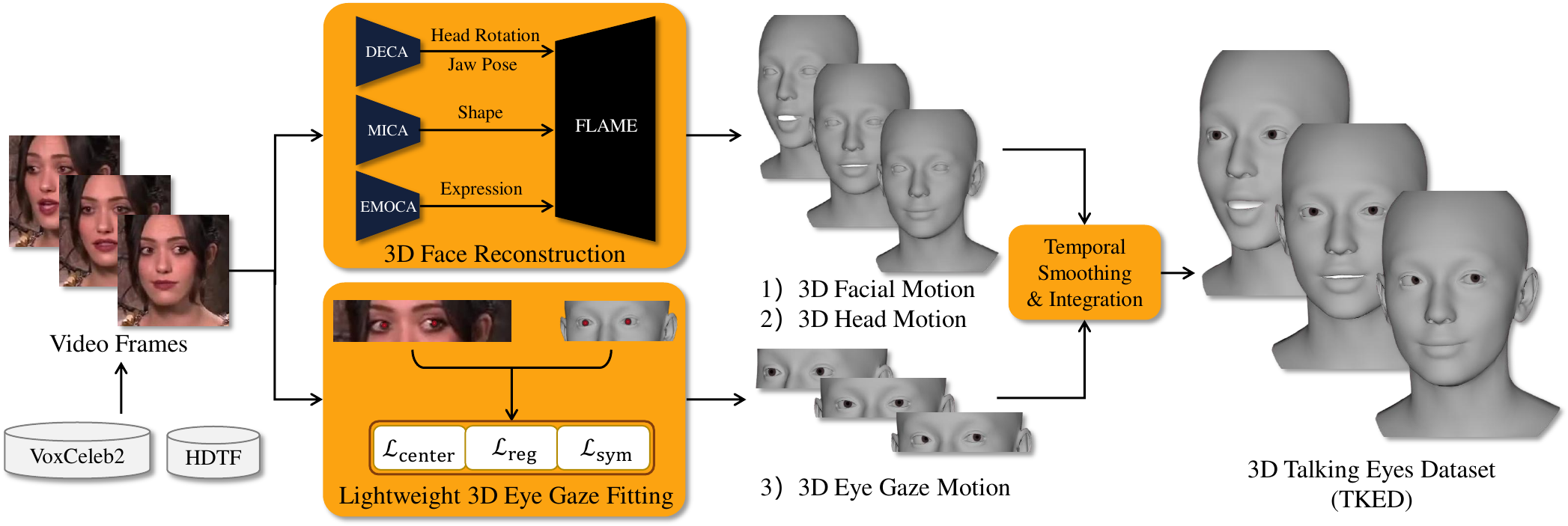}
    \caption{The pipeline of the dataset construction.}
    \label{fig:dataset_pipeline}
\end{figure*}

\subsection{Dataset Construction Pipeline}
\label{sect:dataset_pipeline}
As shown in Fig. \ref{fig:dataset_pipeline}, the pipeline mainly consists of four stages: video pre-processing, 3D face reconstruction, 3D eye gaze fitting and data post-processing.

\subsubsection{Video Pre-processing}
Clear visibility of the subject's eyes in video is paramount for the subsequent 3D eye gaze fitting. Hence, we excluded videos where the eyes are obscured. For instances of poor visibility due to head turns, we performed a preliminary exclusion by detecting significant disparity in the distances from the left and right eye corners to the two sides of the face. Additionally, the videos featuring individuals with obstructions such as sunglasses, which inherently mask the eyes, were excluded by manual screening. After this pre-processing step, the number of the raw video clips was reduced from approximately 20,000 to 6,000.

\subsubsection{3D Face Reconstruction}
We performed frame-by-frame reconstruction on videos to generate the 3D facial motion and 3D head motion. In order to obtain the highest quality possible, we utilized the 3D face reconstruction network proposed in EMOTE \cite{EMOTE2023}, which combines the strengths of four SOTA methods: EMOCA\cite{EMOCA2022}, SPECTRE\cite{LipreadVideo2022}, DECA\cite{DECA2021} and MICA\cite{zielonka2022towards}. All the four SOTA methods use FLAME as 3D face representation. On the output side, DECA outputs the global rotation of head, forming the 3D head motion; MICA outputs the facial shape vector, EMOCA outputs the expression vector and DECA outputs the jaw pose, together forming the 3D facial motion. SPECTRE is only used in network training to enhance lip articulation by lip-reading loss.

\subsubsection{3D Eye Gaze Fitting}
The above 3D face reconstruction network can't generate 3D eye gaze motion. Hence, we propose a lightweight 3D eye gaze fitting method to capture 3D eye gaze from Internet videos. This method is introduced in the next subsection.

\subsubsection{Post-processing}
Post-processing refines the motion data obtained from the previous steps. 
For frames containing blinks, the eye pose was maintained from the last known open-eye state because the iris position can't be reliably detected during a blink. This ensures the continuity and naturalness of the facial animation by preventing unnatural eye movements during blinks. We then applied temporal convolutional smoothing individually to the three types of motion data to reduce jitter and finally combined them in a single FLAME model.



\subsection{LightGazeFit}
\label{sect:LightGazeFit}
The deep-learning-based gaze estimation methods \cite{GazeEstimationSurvey2024} proposed in computer vision provide a number of potential solutions for reconstructing eye gaze from video, but their estimated 3D gaze directions are not always compatible with the reconstructed 3D face mesh. Alternatively, 3D eye gaze fitting methods \cite{wang2016realtime, wang2019realtime, wen2017real} from computer graphics seem to offer a better fit, but their code has not been publicly released.

\begin{figure}
    \centering
    \includegraphics[width=0.48\textwidth]{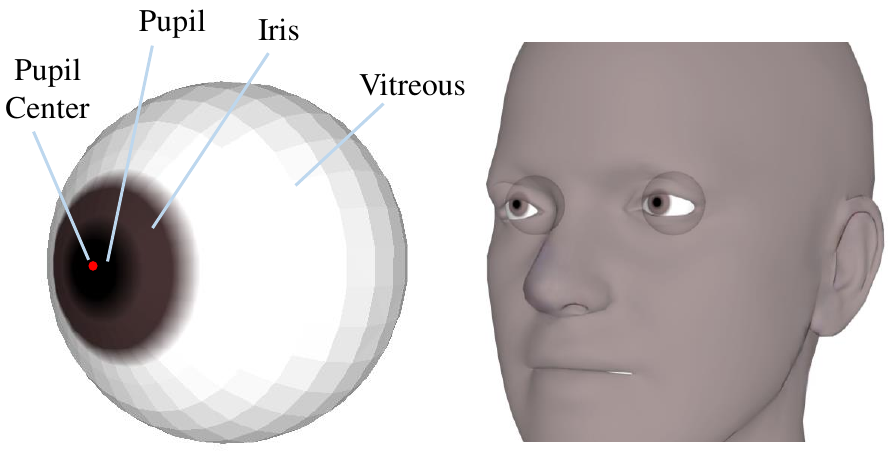}
    \caption{The 3D eyeball model used in our method (left) and its relative position and orientation to the FLAME head model (right). }
    \label{fig:eyeball}
\end{figure}

The proposed LightGazeFit is a lightweight yet effective 3D eye gaze fitting method tailored for processing videos from VoxCeleb2 and HDTF. Existing fitting methods \cite{wang2016realtime, wang2019realtime} heavily rely on accurate segmentation of iris and pupil pixels to formulate mask or edge likelihood. However, due to the relatively poor quality of VoxCeleb2 videos, which are only available in a $224\times224$ resolution, achieving precise segmentation of iris and pupil is nearly impossible. LightGazeFit overcomes this issue by utilizing only the pupil center to form the likelihood, yet it still generates good results across all videos from VoxCeleb2 and HDTF.


The 3D eyeball model used in LightGazeFit comes from FLAME and it's shown in Fig. \ref{fig:eyeball}. We manually annotated the iris vertices, the pupil vertices and the pupil center vertex on the 3D model. Then, we assigned a brown color to the iris vertices and a black color to the pupil vertices.
As the precise segmentation of iris and pupil is not available, LightGazeFit eliminates the need of eyeball calibration \cite{wang2016realtime, wang2019realtime, wen2017real} which estimates the iris and pupil size for each subject. With the iris and pupil size fixed, we found that it performs well in the eyeball rotation estimation.  

The objective function used in LightGazeFit is defined as:
\begin{equation}
\label{eq:final_objective}
\begin{split}
    \mathcal{L} = \mathcal{L}_{\text{center}} + \lambda_1\mathcal{L}_{\text{reg}} +
    \lambda_2\mathcal{L}_{\text{sym}},
\end{split}    
\end{equation}
where $\lambda_1$ and $\lambda_2$ are hyper-parameters and are set to $1e-4$ and $1e-2$ respectively. The pupil center loss ${\mathcal{L}_{\text{center}}}$ is defined as the difference between the predicted 2D pupil center $\mathbf{y}$ and the point $\hat{\mathbf{y}}$ projected from the pupil center vertex on 3D eyeball model to the 2D image plane:
\begin{equation}
    \mathcal{L}_{\text{center}} = ||\hat{\mathbf{y}}-\mathbf{y}||_2^2.
\end{equation}
The 2D pupil center $\mathbf{y}$ in the image was predicted by a neural network \cite{wang2019realtime}. Given the 2D pupil center $\mathbf{y}$, their exists many 3D eyeball rotation solutions to generate $\hat{\mathbf{y}}$. To reduce such ambiguity, we used regularization loss to make the eyeball rotate as little as possible. The regularization loss $\mathcal{L}_{\text{reg}}$ is defined as:
\begin{equation}
    \mathcal{L}_{\text{reg}} = ||\mathbf{p}_{eye}||_2^2,
\end{equation}
where $\mathbf{p}_{eye} \in \mathbb{R}^{6}$ denotes the 3D rotation of two eyeballs. Finally, based on the assumption that left and right eyeball share the similar behaviour, the symmetry loss $\mathcal{L}_{\text{sym}}$ is defined as:
\begin{equation}
    \mathcal{L}_{\text{sym}} = ||\mathbf{p}_{eye\_l}-\mathbf{p}_{eye\_r}||_2^2.
\end{equation}

We used the Adam optimizer with a learning rate of $5e-3$ to solve this non-linear least square optimization problem defined as Eq. \ref{eq:final_objective}.






\section{Method: TalkingEyes}
Our ultimate goal is to collectively synthesize eye gaze motion, eye blinks, head motion and facial motion from speech, under the unified 3D face representation FLAME \cite{FLAME2017}. For facial motion, we directly use Learn2Talk \cite{Learn2Talk2025} as the generator due to its impressive 3D lip-sync performance. Eye gaze motion and head motion are firstly transformed into their respective latent spaces via VQVAE and VAE, and are then jointly generated by a temporal autoregressive model.
We detail the motivation of using VAE and VQVAE in Sect. \ref{sect:motivation_VAEVQVAE}, the pipeline of TalkingEyes in Sect. \ref{sect:method_pipeline}, the two main stages in training in Sect. \ref{sect:discrete_space} and \ref{sect:speech-to-headgaze}, and the synthesis of eye blinks in Sect. \ref{sect:eyeblink}.


\begin{figure*}
    \centering
    \includegraphics[width=0.96\textwidth]{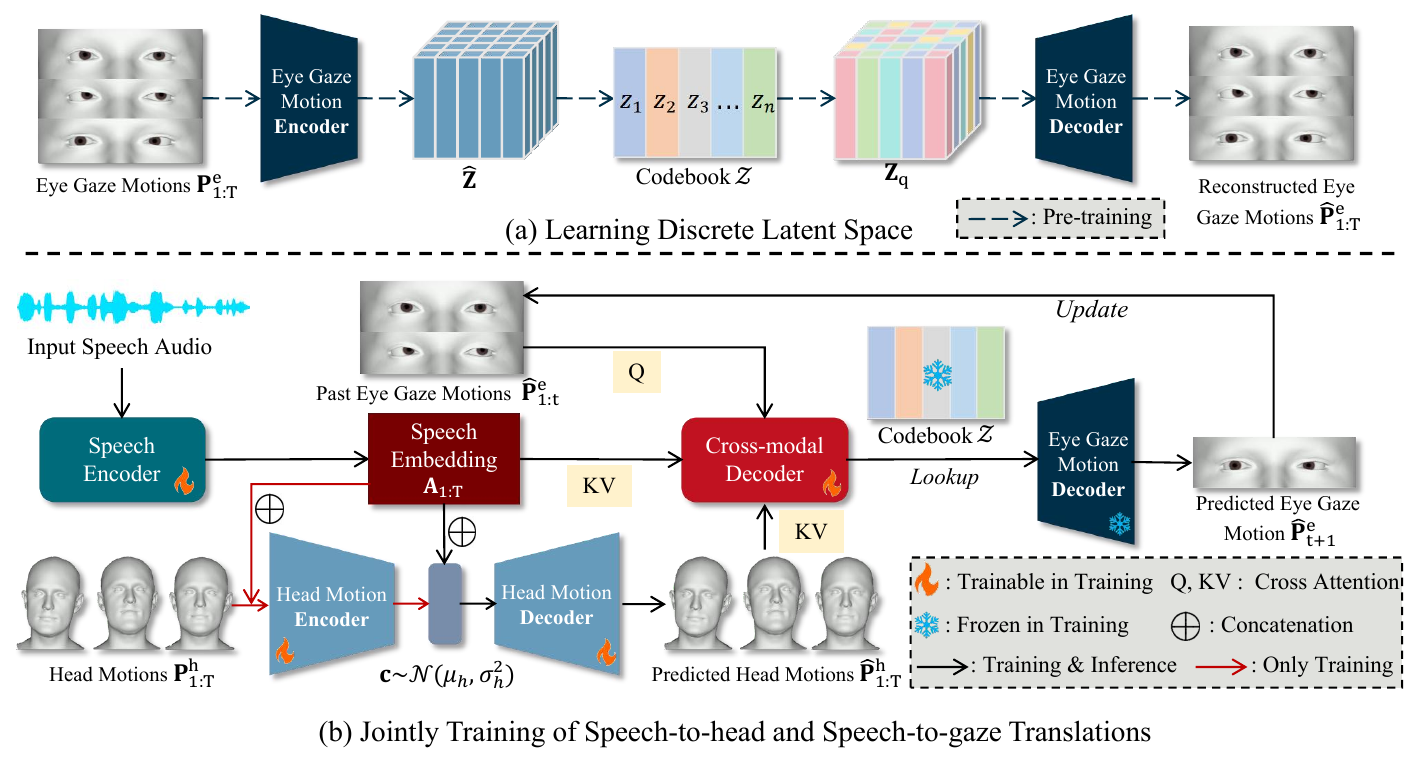}
    \caption{The pipeline of TalkingEyes. The training comprises two stages: (a) learning the discrete latent space for eye gaze motions by pre-training VQVAE, (b) jointly training the VAE-based speech-to-head translation and the VQVAE-based speech-to-gaze translation in an autoregressive manner. In the inference, the ground truth head motions and the head motion encoder in VAE are removed.}
    \label{fig:pipeline}
\end{figure*}

\begin{table}
\centering
\fontsize{8.0}{10.0}\selectfont
\begin{tabular}{l|c|c}
  \hline
  &Diversity &Realism \\  
  \hline
  Head: VAE& high & natural \\  
  Head: VQVAE& extremely high & exaggerated\\  
  Eye Gaze: VAE & low  & nearly stationary \\  
  Eye Gaze: VQVAE& high  & natural \\ 
  \hline
\end{tabular}
\caption{Comparison of different choices of VAE and VQVAE in the latent space learning for head motion and eye gaze motion.}
\label{tab:VAEVQVAE}
\end{table}

\subsection{Latent Spaces Learning by VAE and VQVAE}
\label{sect:motivation_VAEVQVAE}
To model the weak correlation between speech signals and non-verbal cues in conversation, such as head gesture \cite{SadTalker2023}, hand gesture \cite{TalkShow2023, GestureDiffuCLIP2023} and body poses \cite{EMAGE2024}, the generative models VAE \cite{vae2014, vae2016} and VQVAE \cite{VQVAE2017, VQVAE2-2019} are commonly used. These models are capable of learning a compact representation for the non-verbal motion data. Through mapping the speech to the compact latent space, the difficulty in capturing the weak correlation between speech and motion is significantly attenuated and hence promotes the quality of motion synthesis. Following this, TalkingEyes also employ VAE and VQVAE to learn the latent space of motion.

On the other hand, there is a key difference between VAE and VQVAE in terms of motion diversity. VAE tends to output the mean of a Gaussian distribution, thus limiting the diversity in outputs. VQVAE uses discrete codebook to capture a diverse set of latent representations, allowing for a wider range of possible outputs. Since the rotation range of head usually exceeds that of eyeballs, the variation of the head motions is generally higher than that of eye gaze motions. As summarized in Tab. \ref{tab:VAEVQVAE}, VAE is capable of producing head motions with adequate diversity (e.g. PoseVAE \cite{SadTalker2023, Learn2Talk2025}), while VQVAE may sometimes produce excessive diverse and exaggerated head motions. In contrast, VAE struggles to produce sufficiently diverse eye gaze motions, while VQVAE effectively achieve this. To support our claims, we have conducted quantitative experiments which are introduced in Sect. \ref{sect:ablation}. To conclude, we use VAE in the latent space learning for head motions, and use VQVAE in that for eye gaze motions.


\subsection{Pipeline}
\label{sect:method_pipeline}
The pipeline of TalkingEyes is shown in Fig. \ref{fig:pipeline}, featuring the translations from speech signals to 3D eye gaze motions $\mathbf{\hat{P}}^e_{1:T}\in \mathbb{R}^{T\times6}$ (left and right eyes) and 3D head motions $\mathbf{\hat{P}}^h_{1:T}\in \mathbb{R}^{T\times3}$ in one framework, where $T$ denotes the number of frames. 

The training of TalkingEyes comprises two main stages: firstly learn a discrete latent space for eye gaze motions by pre-training VQVAE, and then jointly train the VAE-based speech-to-head translation and the VQVAE-based speech-to-gaze translation in an autoregressive manner. 
In the first stage, we pre-train a VQVAE to model the latent space of eye gaze motions as a discrete codebook $\mathcal{Z}$. The codebook is learned by self-reconstruction over the ground truth motions $\mathbf{P}^e_{1:T}$. 
In the second stage, a conditional VAE is employed to reconstruct the head motions $\mathbf{\hat{P}}^h_{1:T}$ from the ground truth motions $\mathbf{P}^h_{1:T}$. This motion reconstruction is conditioned on the speech embedding $\mathbf{A}_{1:T}$ that is extracted by the speech encoder (wav2vec2.0 \cite{wav2vec2020}) from the input speech. Then the predicted head motions $\mathbf{\hat{P}}^h_{1:T}$ and the previously predicted eye gaze motions $\mathbf{\hat{P}}^e_{1:t}$ are used as conditions in the mapping of the speech embedding $\mathbf{A}_{1:T}$ to the target motion codes. This mapping is accomplished through multi-modal alignment and codebook $\mathcal{Z}$ lookup. The motion codes are further decoded into the eye gaze motion $\mathbf{\hat{P}}^e_{t+1}$ at current frame $t+1$, which will serve as one frame of past motions in the next round of autoregression.

In the inference, the ground truth head motions $\mathbf{P}^h_{1:T}$ and the head motion encoder in VAE are not used. The head latent code $\mathbf{c}$ in VAE is randomly sampled from a standard Gaussian.




\subsection{Discrete Latent Space of Eye Gaze Motion}
\label{sect:discrete_space}
We construct a discrete codebook $\mathcal{Z}= 
\{ {\mathbf{z}_k}\in \mathbb{R}^{C} \}_{k=1}^N$ to form the discrete latent space, thus allowing any frame in a eye gaze motion sequence to be represented by a codebook item $\mathbf{z}_k$. The Transformer-based VQVAE model that consists of a motion encoder $E_{\text{VQ}}$, a motion decoder $D_{\text{VQ}}$ and a codebook $\mathcal{Z}$, is pre-trained under the self-reconstruction of ground truth motions $\mathbf{P}^e_{1:T}$. As shown in Fig. \ref{fig:pipeline}, the motions $\mathbf{P}^e_{1:T}$ is firstly encoded into a temporal feature vector $\hat{\mathbf{Z}}$. Then, $\hat{\mathbf{Z}}$ is quantized to the feature vector ${\mathbf{Z}_q}$ via a element-wise quantization function $Q(\cdot)$ that maps each item in $\hat{\mathbf{Z}}$ to its nearest entry in codebook $\mathcal{Z}$: 
\begin{equation}
\label{eq:lookup}
\mathbf{Z}_{q}=Q(\mathbf{\hat Z}):=\underset{\mathbf{z}_{k}\in \mathcal{Z}}{\arg \min }\|\mathbf{\hat z}_{t}-\mathbf{z}_{k}\|_{2}.
\end{equation}
${\mathbf{Z}_q}$ is finally decoded into the reconstructed motions $\mathbf{\hat{P}}^e_{1:T}$ via:
\begin{equation}
\label{eq:DVQEVQ}
\mathbf{\hat{P}}^e_{1:T}=D_{\text{VQ}}(\mathbf{Z}_q) = D_{\text{VQ}}(Q (E_{\text{VQ}}(\mathbf{P}^e_{1:T})))
\end{equation}

Similar to \cite{VQVAE2017, CodeTalker2023}, the training losses of VQVAE include a motion-level loss and two intermediate code-level losses:
\begin{equation}
    \mathcal{L} = \|\mathbf{P}^e_{1:T} - \mathbf{\hat{P}}^e_{1:T}\|_2^2 \\
    + \|sg(\mathbf{\hat{Z}}) - \mathbf{Z}_q\|_2^2 + \beta \| \mathbf{\hat{Z}} - sg(\mathbf{Z}_q)\|_2^2,
\end{equation}
where the first term is a motion reconstruction loss, the latter two terms
update the codebook by reducing the distance between codebook $\mathcal{Z}$ and embedded feature $\hat{\mathbf{Z}}$, $sg(\cdot)$ stands for a stop-gradient operation and $\beta$ is a weighting factor.

\subsection{Speech-to-head and Speech-to-gaze Translations}
\label{sect:speech-to-headgaze}
We adopt a conditional VAE model to model the translation from speech to head motions. The head motion encoder $E_{\text{VAE}}$ accepts ground truth motions $\mathbf{P}^h_{1:T}$ concatenated with speech embedding $\mathbf{A}_{1:T}$ as input. The encoder $E_{\text{VAE}}$ outputs the continuous latent code $\mathbf{c}$ that adheres a Gaussian distribution $\mathcal{N}(\mu_h,\sigma_h^2)$ with the learned mean $\mu_h$ and learned variance $\sigma_h^2$. The latent code $\mathbf{c}$ is further concatenated with $\mathbf{A}_{1:T}$, and then be inputted to the head motion decoder $D_{\text{VAE}}$ to predict head motions as $\mathbf{\hat{P}}^h_{1:T}$. All the $T$-frames of motions in  $\mathbf{\hat{P}}^h_{1:T}$ are generated by once reconstruction, which can be formulated as:
\begin{equation}
\mathbf{\hat{P}}^h_{1:T}=D_{\text{VAE}}(\textbf{c}, \mathbf{A}_{1:T})=D_{\text{VAE}}(E_{\text{VAE}}(\mathbf{P}^h_{1:T}, \mathbf{A}_{1:T}), \mathbf{A}_{1:T}).
\end{equation}

Subsequent to the VAE model, a VQVAE-based autoregressive model is cascaded for the eye gaze motion generation. The predicted head motions $\mathbf{\hat{P}}^h_{1:T}$ is used as condition to guide the translation from speech to eye gaze motions.
Specifically, we adopt a Transformer cross-modal decoder $D_{\text{cross-modal}}$ to algin three different modalities, namely speech audio, head motions and eye gaze motions. $D_{\text{cross-modal}}$ is equipped with causal self-attention to learn the dependencies between each frame in the context of $t$-frames of past eye gaze motions $\mathbf{\hat{P}}^e_{1:t}$. Additionally, $D_{\text{cross-modal}}$ is  equipped with cross-modal attention to align the past eye gaze motions $\mathbf{\hat{P}}^e_{1:t}$ respectively with the head motions $\mathbf{\hat{P}}^h_{1:T}$ and the speech embedding $\mathbf{A}_{1:T}$. We set the cross-attention keys (K) and values (V) as the concatenation of the predicted head motions embedding and the speech embedding, and set the queries (Q) as the past eye gaze motions embedding. The cross-modal decoder $D_{\text{cross-modal}}$ outputs the features $\hat{\mathbf{Z}}^{1:t}$ as:
\begin{equation}
    \hat{\mathbf{Z}}^{1:t} = D_{\text{cross-modal}}(\mathbf{A}_{1:T}, \mathcal{P}^e_{\text{emb}}(\mathbf{\hat{P}}^e_{1:t}), \mathcal{P}^h_{\text{emb}}(\mathbf{\hat{P}}^h_{1:T})),
\end{equation}
where $\mathcal{P}^e_{\text{emb}}$ and $\mathcal{P}^h_{\text{emb}}$ are linear projection layers to extract eye gaze motions embedding and head motions embedding respectively. 
$\hat{\mathbf{Z}}^{1:t}$ is further quantized into $\mathbf{Z}^{1:t}_q$ via Eq. \ref{eq:lookup} and decoded by the pre-trained VQVAE decoder into $\mathbf{\hat{P}}^e_{t+1}$ via Eq. \ref{eq:DVQEVQ}, which can be formulated as:
\begin{equation}
\mathbf{\hat{P}}^e_{t+1}=D_{\text{VQ}}(\mathbf{Z}^{1:t}_q)=D_{\text{VQ}}(Q(\hat{\mathbf{Z}}^{1:t})).
\end{equation}
The newly predicted motion $\mathbf{\hat{P}}^e_{t+1}$ is used to update the past motions, in preparation for the next prediction.




\textbf{Training Losses.}  We train the head motion encoder $E_{\text{VAE}}$, the head motion decoder $D_{\text{VAE}}$, the cross-modal decoder $D_{\text{cross-modal}}$, the projection layers $\mathcal{P}^e_{\text{emb}}, \mathcal{P}^h_{\text{emb}}$ and part of the speech encoder, while keeping the codebook $\mathcal{Z}$ and eye gaze motion decoder $D_{\text{VQ}}$ frozen. The training losses consist of four terms: reconstruction loss ${\mathcal{L}_{\text{rec}}}$, velocity loss ${\mathcal{L}_{\text{vel}}}$, KL-divergence loss ${\mathcal{L}_{\text{kl}}}$ and feature regularity loss ${\mathcal{L}_{\text{reg}}}$. 

The reconstruction loss ${\mathcal{L}_{\text{rec}}}$ measures the difference between predicted motions and ground truth motions for both head and eye gaze:
\begin{equation}
    \mathcal{L}_{\text{rec}} = ||\hat{\mathbf{P}}^h_{1:T}-\mathbf{P}^h_{1:T}||_2^2 + ||\hat{\mathbf{P}}^e_{1:T}-\mathbf{P}^e_{1:T}||_2^2.
\end{equation}

The velocity loss ${\mathcal{L}_{\text{vel}}}$ measures the discrepancy between the first-order derivative of predicted motions and that of ground truth motions for both head and eye gaze:
\begin{equation}
\label{eq:velocity_loss}
\begin{split}
\mathcal{L}_{\text{vel}} = \sum_{t=1}^{T-1}||(\mathbf{\hat{P}}^h_{t+1}-\mathbf{\hat{P}}^h_t)-(\mathbf{P}^h_{t+1}-\mathbf{P}^h_t)||_2^2 
\\ + \sum_{t=1}^{T-1}||(\mathbf{\hat{P}}^e_{t+1}-\mathbf{\hat{P}}^e_t)-(\mathbf{P}^e_{t+1}-\mathbf{P}^e_t)||_2^2.
\end{split}
\end{equation}

The KL-divergence loss ${\mathcal{L}_{\text{kl}}}$ forces the Gaussian distribution of the predicted head motions $\mathbf{P}^h_{1:T} \sim \mathcal{N}(\mu_h,\sigma_h^2)$ close to a standard Gaussian $\mathcal{N}(0,1)$. ${\mathcal{L}_{\text{kl}}}$ has a simplified form \cite{vae2014}:
\begin{equation}
    \mathcal{L}_{\text{kl}} = - \frac{1}{2} (1 + \log{\sigma_h^2} - \sigma_h^2 - \mu_h^2).
\end{equation}

The feature regularity loss ${\mathcal{L}_{\text{reg}}}$ measures the deviation between the predicted eye gaze motion feature $\mathbf{\hat{Z}}^{1:T}$ and the quantized feature $\mathbf{Z}^{1:T}_q$ from codebook:
\begin{equation}
\label{eq:regularity_loss}
    \mathcal{L}_{\text{reg}} = ||\mathbf{\hat{Z}}^{1:T} - sg(\mathbf{Z}^{1:T}_q) ||_2^2.
\end{equation}

The final training loss is formulated as follows:
\begin{equation}
    \mathcal{L} = \mathcal{L}_\text{rec} + \mathcal{L}_\text{vel} + \lambda_1\mathcal{L}_\text{kl} + \mathcal{L}_\text{reg},
\end{equation}
where $\lambda_1$ is set to 1e-4.

\subsection{Eye Blink}
\label{sect:eyeblink}
To generate more vivid animation, we have also implemented the synthesis of eye blinks. Blinking is primarily influenced by personal habits, environmental factors and other conditions.
Liu et al. \cite{liu2024emoface} counted the number of blinks in their dataset and found that the frequency of human blinking, measured in blinks per minute, typically follows a log-normal distribution. Following their work, we conducted a statistical analysis on our dataset TKED to derive a suitable blink frequency that well aligns with our eye gaze animation. We counted the number of blinks in each video in TKED and then converted these counts into blinks per minute based on the video's duration. For blink detection, we calculated the Eye Aspect Ratio (EAR) \cite{cech2016real}, which is defined as the ratio of the eye's height to its width. When the eyes are closed, the EAR approaches zero. After statistical analysis, we found that the blinks per minute follow a Gaussian distribution with a mean of 40.10 and a deviation of 15.42.
When generating animations, we first sample the number of blinks per minute from the distribution and then convert that into blink intervals. Subsequently, we adjust the expression parameters within a 5-frame window for each blink to create the blinking action.

\section{Experiments}
\label{sect:experiments}
\subsection{Training Details}
VQVAE is pre-trained by Adam optimizer for 100 epochs with a learning rate $1e-4$. The pre-training takes about $1$ hour on a NVIDIA RTX 4090 GPU. Then the VAE and the autoregressive model are trained by Adam optimizer for 100 epochs with a learning rate $1e-4$. The batch size is set to 1. The training takes about $5$ hours on the same GPU.

\subsection{Experimental Setup}

\subsubsection{Dataset} Our method and the baselines are trained and evaluated on TKED, following its training (TKED-Train), validation (TKED-Val) and testing (TKED-Test) splits.

\subsubsection{Baselines} Since the recent speech-driven 3D eye gaze animation methods are scarce and early computer animation methods \cite{HeadEyeAnimation2007, VirtualCharacterSCA2013, ConversationalAgents2012, LiveSpeehEyeTVCG2012, EyeMotion2019} targeted at this task have not released their codes or datasets, we implemented four 3D eye gaze motion generation methods as the baselines for quantitative comparison. 
\begin{itemize}
\item[--] \emph{Mean.} For any input audio in TKED-Test, this method always outputs the average value of eye rotation vectors in TKED-Train. As a result, the generated eye gaze by this method is always stationary.
\item[--] \emph{Random.} For an input audio in TKED-Test, this method randomly selects a eye gaze motion sequence in TKED-Train that does not correspond to the input audio as prediction. This method can produce sufficient diverse motions for the same input.
\item[--] \emph{Similar Sample.} For an input audio in TKED-Test, this method firstly extracts features from the input, then identifies the audio in TKED-Train with the most similar features to the input, and uses the corresponding eye gaze motion sequence as prediction. This method seems more reasonable than Mean and Random.
\item[--] \emph{VAE.} Similar to the speech-to-head translation (Sect. \ref{sect:speech-to-headgaze}), this method uses conditional VAE model to predict eye gaze motions from speech. Through concatenation, both the head motions and the speech embedding are used as condition.
\end{itemize}

\subsubsection{Evaluation Metrics} We assess the quality of the generated 3D eye gaze motions by different methods in the following aspects.
\begin{itemize}
\item[--] \emph{Motion Diversity.} Because we model the motion generation as a non-deterministic task, we manage to measure how much the generated motions diversify with the same audio input. Specifically, we run a motion generation method multiple times for the same input audio, and obtain a set of plausible motions. The average $L_2$ distance of all paired motions in the sets represents the diversity score.
\item[--] \emph{Audio-Motion Correlation.} Although audio weakly correlates with eye gaze motion, we still explore to measure the degree of association between them. We use contrast learning to compute the similarity between audio and motion features. A binary classification network is trained with positive and negative samples. The positive samples consist of audio clips and their well synchronized motion clips, while the negative samples comprise mismatched audio and motion clips. Given a pair of audio and motion, the correlation is measured by the cosine similarity between the two features extracted by the classification network.
\item[--] \emph{Motion Discrepancy.} The discrepancy is measured by the $L_2$ distance between the predicted motions and the ground truth motions. As this metric doesn't reflect the intrinsic correlation between audio and the generated diverse motions, we list it in the evaluation for reference only. 
\end{itemize}

\subsection{Evaluation of TalkingEyes}

\subsubsection{Quantitative Comparison} 
Tab. \ref{tab:quantitative_evaluation} shows the quantitative evaluation results over TKED-Test. Our TalkingEyes outperforms other methods in terms of motion diversity and audio-motion correlation. TalkingEyes is capable of generating more diverse motions than the ``Random" selection, thanks to VQVAE's ability to learn a structured and clustered representation from the original data. Additionally, TalkingEyes achieves the highest score in audio-motion correlation, indicating that mapping the audio signal to the discrete motion space is a more effective and robust way for modeling the weak relationship between audio and eye gaze motion compared to other approaches, such as searching for similar samples in the audio feature space. When it comes to $L_2$ distance, TalkingEyes slightly compromises accuracy in matching the ground truth, which is a common trade-off in non-deterministic tasks \cite{PICCVPR2019, ICTPAMI2024}. It is worth noting that ``VAE" achieves very similar performance to that of ``Mean" with respect to the three evaluation metrics. This further supports our motivation in the choice of VAE or VQVAE, as discussed in Sect. \ref{sect:motivation_VAEVQVAE}: VAE tends to output the mean of the Gaussian constructed on the eye gaze motions.

\begin{table}
\centering
\fontsize{8}{10}\selectfont
\begin{tabular}{c|ccc}
	  \hline
	Method & Div. ↑ & Corr. ↑ & $L_2$ ↓ \cr
	  \hline
	Mean & 0.0 & 0.45685 & \textbf{0.17162} \cr
	Random & 0.23383 & 0.49117 & 0.23989\cr
	Similar Sample & 0.06170 & 0.57849 & 0.23479\cr
	VAE & 0.06655 & 0.45796 & 0.17887 \cr
        \hline
	TalkingEyes (ours) & \textbf{0.26398} & \textbf{0.60435} & 0.26570 \cr
	  \hline
\end{tabular}
\caption{Methods comparison. The quantitative evaluation
results over TKED-Test are reported. The
best score in each metric is marked with bold.}
\label{tab:quantitative_evaluation}
\end{table}

\begin{table}
\centering
\fontsize{8}{10}\selectfont
\begin{tabular}{l|c}
	\hline
    {Method A vs. Method B}  & Preference Rate\cr
	\hline
    Ours vs. Mean & 81 (1.68e-4) \cr
    Ours vs. Random & 80 (1.34e-4) \cr
    Ours vs. Similar Sample & 81 (1.05e-4) \cr
    Ours vs. VAE & 68 (3.28e-2) \cr
    Ours vs. Ground Truth & 47 (7.11e-1) \cr
    Random vs. Ground Truth & 34 (9.77e-3) \cr
    \hline
\end{tabular}
\caption{User study results. The percentage of answers where Method A is preferred over Method B is listed. The numbers in the brackets denote p-value.} 
\label{tab:userstudy}
\end{table}

\subsubsection{Perceptual User Study} For more comprehensive evaluation, we have conducted a perceptual user study to evaluate the realism of the animated 3D eye gaze. We employed an A/B test, presenting comparisons in a random order to evaluate our TalkingEyes method against four baseline methods and ground truth recordings.
In each comparison, the participants were presented with two video clips with audio and asked: ``Which recording exhibits more natural eyeball movements with fewer artifacts, such as speech-unrelated motions, exaggerated movements, or static poses?" They responded by selecting either A or B. For the study, we randomly selected 20 audio clips from TKED-Test, and used them to generate animations for six kinds of A/B comparisons (5 audios $\times$ 6 comparisons). 
20 participants took part in the study, finally yielding 600 entries.
We calculated the preference rate for Method A by determining the percentage of times it was chosen over Method B in a pairwise comparison. The results of this A/B testing are tabulated in Tab. \ref{tab:userstudy}. We also conducted Wilcoxon test in the study, in which the p-value less than 0.05 indicates a statistically significant difference between Method A and B.

As clearly shown in Tab. \ref{tab:userstudy}, participants preferred our method over all the four baselines. However, they still favored the ground truth recording over our method, which validates our experimental setup. The last comparison (``Random vs. Ground Truth") confirms that participants can perceive the correspondence between speech and eye gaze motion, as the ``Random" baseline pairs ground truth eye gaze motion with unrelated audio.

 \begin{figure}
    \centering
    \includegraphics[width=0.48\textwidth]{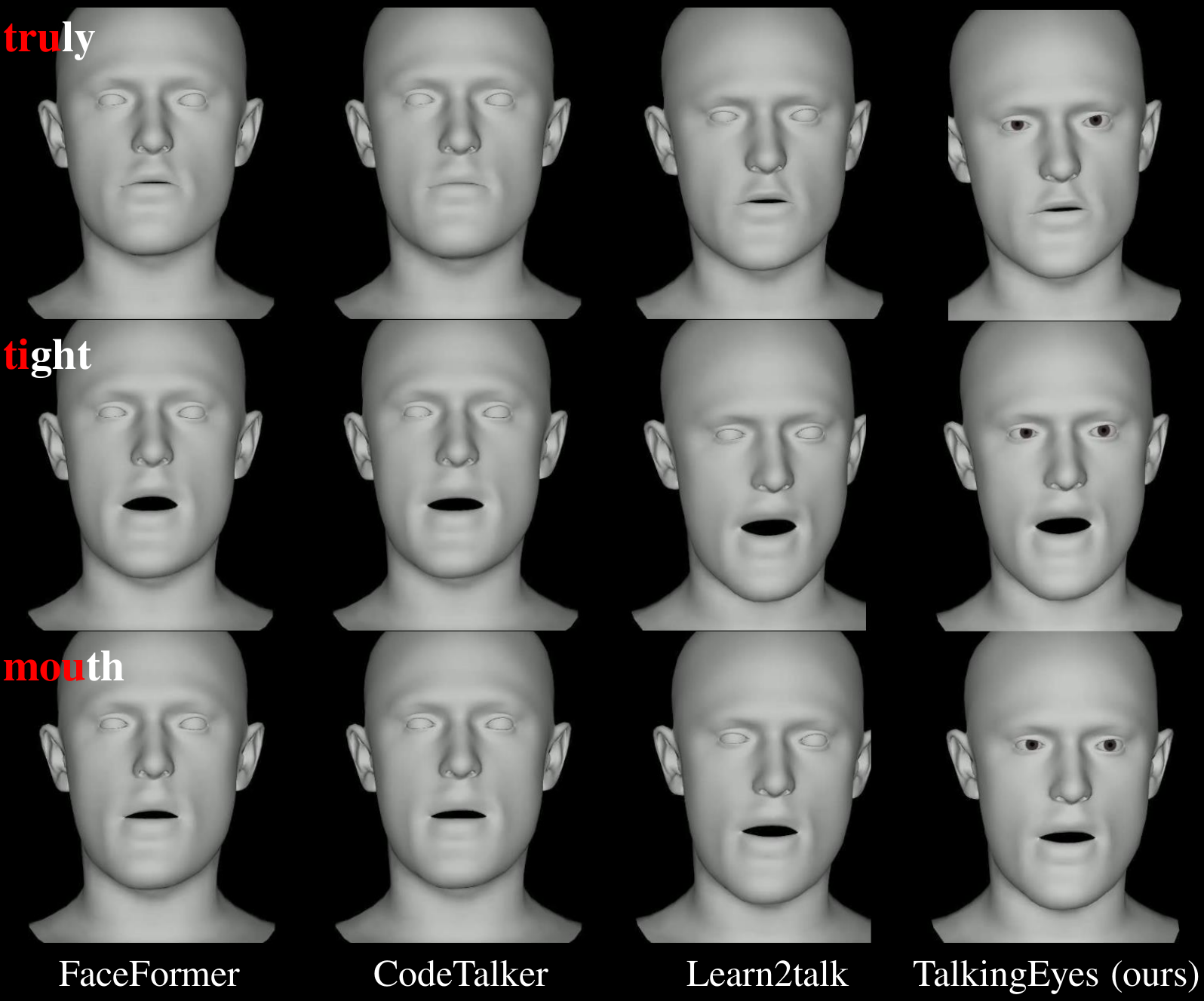}
    \caption{Visual comparisons of sampled frames of 3D facial animations generated by FaceFormer \cite{FaceFormer2022}, CodeTalker \cite{CodeTalker2023}, Learn2Talk \cite{Learn2Talk2025} and our method.}
    \label{fig:main_compare}
\end{figure}

\subsubsection{Qualitative Comparison} 
Visual comparisons between our method and the four baselines are provided in the supplementary video. The comparison results demonstrate that, ``Mean" can only generate stationary eye gaze motion, ``Random" and ``Similar Sample" tend to produce disorganized and speech-unrelated eyeball movement, and ``VAE" yields nearly stationary motion. In contrast, our method produces natural eye gaze motion in tune with the speech.
We also conduct visual comparisons with three recent speech-driven 3D facial animation methods in the supplementary video, namely FaceFormer \cite{FaceFormer2022}, CodeTalker \cite{CodeTalker2023} and Learn2Talk \cite{Learn2Talk2025}. In Fig. \ref{fig:main_compare}, we illustrate three typical frames of synthesized facial animations that speak at specific syllables. It's obvious that our method generates the most vivid animation with the animated eye gaze that is synchronized with the speech. Finally, the supplementary video provides a visual comparison with an early speech-driven eye gaze animation method \cite{LiveSpeehEyeTVCG2012}.

\begin{figure*}
    \centering
    \includegraphics[width=1.0\textwidth]{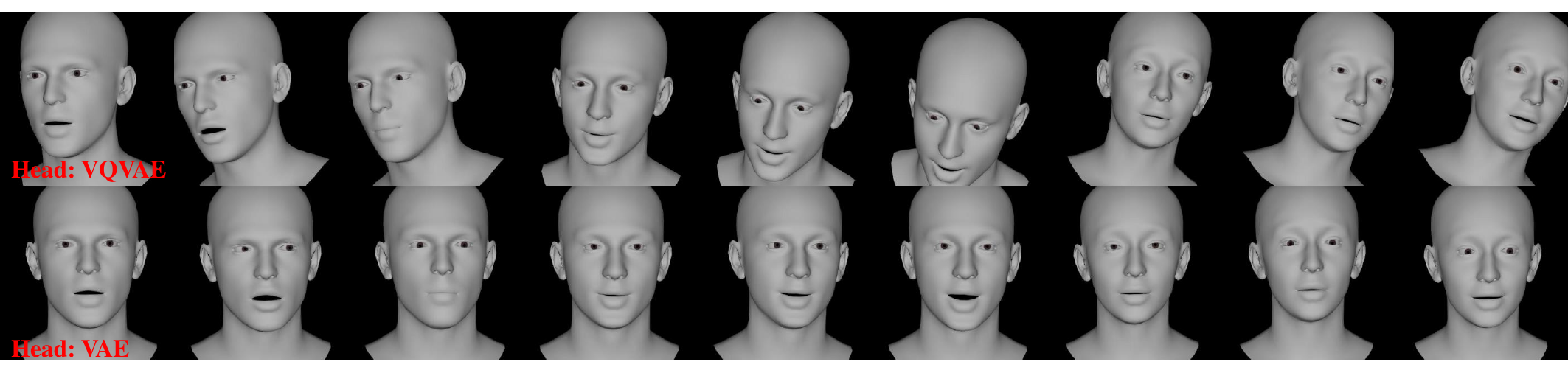}
    \caption{Visual comparison of sampled frames of 3D facial animations generated by using VQVAE and VAE in the latent sapce learning for head motion.}
    \label{fig:vqvae_vae_compare}
\end{figure*}

 \begin{table*}
\centering
\fontsize{8}{10}\selectfont
\setlength{\tabcolsep}{4pt} 
\begin{tabular}{c|cccccc}
	\hline
	Method & $\text{Div.}^h$ ↑ & $\text{Div.}^e$ ↑ & $\text{Corr.}^h$ ↑ & $\text{Corr.}^e$ ↑ & $L_2^h$ ↓ & $L_2^e$ ↓ \cr
	\hline
     Head: VAE, Eye Gaze: VAE & 0.09645 & 0.06655 & \textbf{0.52148} & 0.45796 & 0.23699 & \textbf{0.17887} \cr
     Head: VAE, Eye Gaze: VQVAE (ours) & 0.05143 & \textbf{0.26398} & 0.51913 & \textbf{0.60435} & \textbf{0.22731} & 0.26570 \cr
     Head: VQVAE, Eye Gaze: VQVAE  & \textbf{0.29852} & 0.25439 & 0.45164 & 0.59900 & 0.40460 & 0.26225 \cr
	\hline
\end{tabular}
\caption{Ablation study on the different choices of VAE and VQVAE. The quantitative evaluation results over TKED-Test are reported. The best score in each metric is marked with bold.}
\label{tab:ablation_VAEVQ}
\end{table*}

\subsection{Ablation Analysis}
\label{sect:ablation}
\subsubsection{Different choices of VAE and VQVAE}
We have conducted an ablation study on the different choices of VAE and VQVAE in the latent space learning for head motion and eye gaze motion. The quantitative evaluation results are reported in Tab. \ref{tab:ablation_VAEVQ}. We use the same three evaluation metrics to access the quality of the generated 3D head motions. It could be observed that: 1) compared to eye gaze motions generated by VAE (1st row in Tab. \ref{tab:ablation_VAEVQ}), those generated by VQVAE (2nd row) exhibit higher motion diversity while maintaining a robust correlation with the audio; 2) compared to head motions generated by VAE (2nd row), those generated by VQVAE (3rd row) achieve higher motion diversity but show a lower correlation with the audio. Notwithstanding high diversity, the head motions produced by VQVAE appear exaggerated and unnatural, as shown in Fig. \ref{fig:vqvae_vae_compare}.

\subsubsection{Items in Training Losses}
We have also conducted an ablation study on the two terms of the training losses, namely the velocity loss ${\mathcal{L}_{\text{vel}}}$ defined as Eq. \ref{eq:velocity_loss} and the feature regularity loss ${\mathcal{L}_{\text{reg}}}$ defined as Eq. \ref{eq:regularity_loss}. The other two terms in training losses, including the reconstruction loss ${\mathcal{L}_{\text{rec}}}$ and the KL-divergence loss ${\mathcal{L}_{\text{kl}}}$, are essential losses that can't be removed, as removing them would result in failed training. The quantitative evaluation results are reported in Tab. \ref{tab:ablation_loss}. 

Compared with the full model, removing  ${\mathcal{L}_{\text{vel}}}$ leads to the degeneration in almost all metrics. The artifacts such as motion jitter may arise in the generated long motion sequences. Without the regularization loss ${\mathcal{L}_{\text{reg}}}$ in the quantized features, it's difficult for the training to converge, resulting in the production of stationary 3D eye gaze motions, where the diversity score is zero. 

\begin{table}
\centering
\fontsize{8}{10}\selectfont
\setlength{\tabcolsep}{4pt} 
\begin{tabular}{c|cccccc}
	\hline
	Method & $\text{Div.}^h$ ↑ & $\text{Div.}^e$ ↑ & $\text{Corr.}^h$ ↑ & $\text{Corr.}^e$ ↑ & $L_2^h$ ↓ & $L_2^e$ ↓ \cr
	\hline
     w/o ${\mathcal{L}_{\text{vel}}}$ & 0.05086 & 0.24881 & 0.48854 & 0.59182 & 0.22873 & 0.26941 \cr
     w/o ${\mathcal{L}_{\text{reg}}}$ & 0.00434 & 0.0 & 0.35083 & 0.45581 & 0.48944 & \textbf{0.19682} \cr
     Full model & \textbf{0.05143}  & \textbf{0.26398} & \textbf{0.51913} & \textbf{0.60435} & \textbf{0.22731} & 0.26570 \cr
	\hline
\end{tabular}
\caption{Ablation study on ${\mathcal{L}_{\text{vel}}}$ and ${\mathcal{L}_{\text{reg}}}$. The quantitative evaluation results over TKED-Test are reported. The best score in each metric is marked with bold.}
\label{tab:ablation_loss}
\end{table}

\begin{table}
\centering
\fontsize{8}{10}\selectfont
\setlength{\tabcolsep}{4pt} 
\begin{tabular}{l|cccccc}
	\hline
	Num.  & $\text{Div.}^h$ ↑ & $\text{Div.}^e$ ↑ & $\text{Corr.}^h$ ↑ & $\text{Corr.}^e$ ↑ & $L_2^h$ ↓ & $L_2^e$ ↓ \cr
	\hline
     $N=2048$ & 0.05119 & 0.26278 & \textbf{0.54150} & 0.60268 & 0.22990 & 0.27397 \cr
     $N=1024$ & \textbf{0.05143}  & 0.26398 & 0.51913 & \textbf{0.60435} & \textbf{0.22731} & \textbf{0.26570} \cr
     $N=512$ & 0.04966 & 0.27578 & 0.53699 & 0.59932 & 0.22959 & 0.27345  \cr
     $N=256$ & 0.05027 & \textbf{0.31291} & 0.52527 & 0.58959 & 0.22762 & 0.32000  \cr
	\hline
\end{tabular}
\caption{Ablation study on the number of codebook items. The quantitative evaluation results over TKED-Test are reported. The best score in each metric is marked with bold.}
\label{tab:ablation_codebook}
\end{table}

\subsubsection{Number of Codebook Items}
The quantitative study on the number of items in the codebook is reported in Tab. \ref{tab:ablation_codebook}. With the number of codebook items $N$ decreasing, the diversity of 3D eye motions increases, while the correlation with audio decreases. This phenomenon may be due to the more clustered representation of eye motions, which introduces diversity but diminishes the feature similarity with audio. To balance the diversity and the correlation with audio, we set $N$ to 1024.

\begin{figure}
    \centering
    \includegraphics[width=0.48\textwidth]{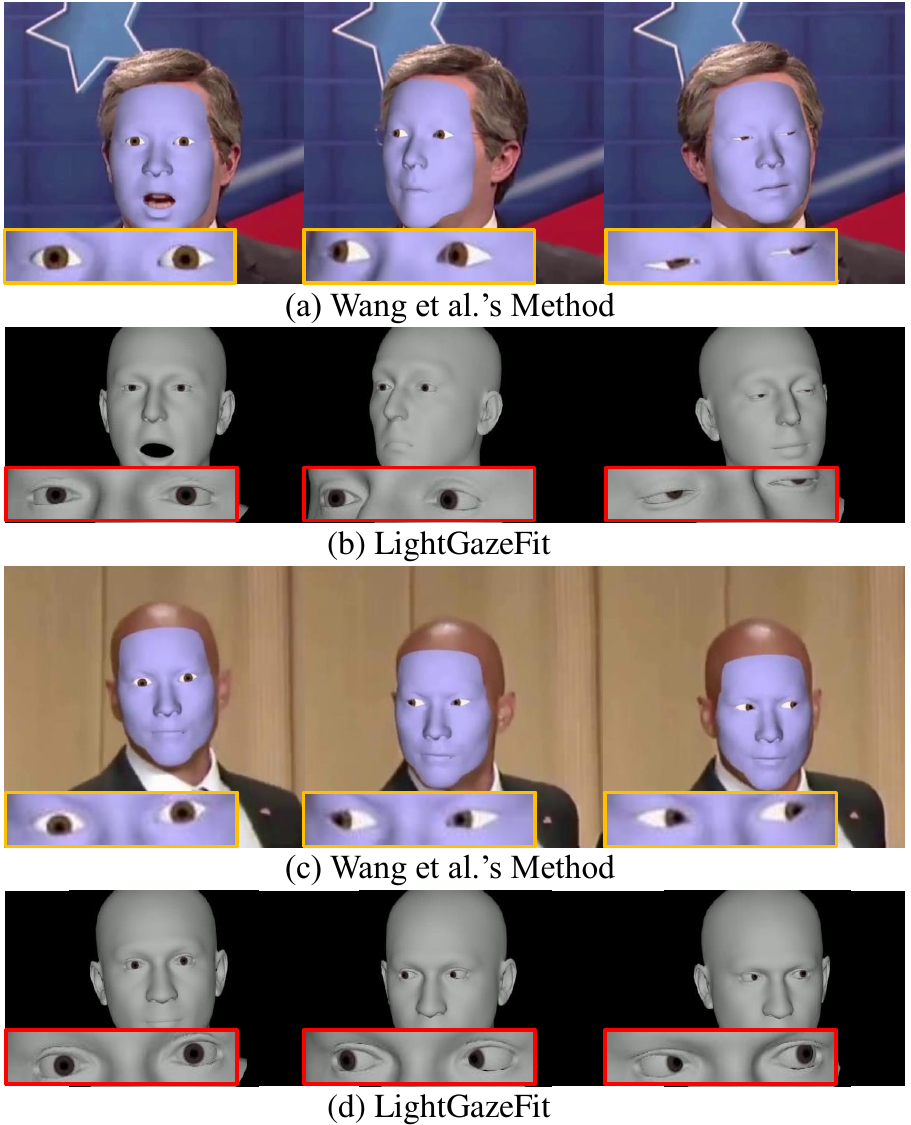}
    \caption{Visual comparisons of 3D eye gaze motions and 3D facial motions reconstructed from input video by using Wang et al.'s method \cite{wang2016realtime} and our method.}
    \label{fig:gaze_compare2}
\end{figure}

\subsection{Evaluation of LightGazeFit}
We compare our LightGazeFit with a 3D eye gaze fitting method, Wang et al.'s method \cite{wang2016realtime}, and a deep-learning-based gaze estimation method, Gaze360 \cite{kellnhofer2019gaze360}. 

\subsubsection{Comparison with Wang et al.'s method}
As shown in Fig. \ref{fig:gaze_compare2}, our method exhibits competitive performance in 3D eye gaze reconstruction, matching that of Wang et al.'s method \cite{wang2016realtime}, across both front view and side views. Our method employs only the pupil center as the fitting constraint, omitting the need for precise segmentation of the iris and pupil, as required by Wang et al.'s method. In this qualitative comparison, the input videos and the 3D reconstruction results of Wang et al.'s method are fetched from the supplementary video in \cite{wang2016realtime}, since the code of Wang et al.'s method has not been publicly released yet.

\subsubsection{Comparison with Gaze360}
We quantitatively compare our method with Gaze360 \cite{kellnhofer2019gaze360}, which trains a deep neural network to predict 3D eye gaze direction from an input facial image. The results are presented in Tab. \ref{tab:quantitative_evaluation_gaze}. 
To construct the evaluation dataset, we randomly selected 30 high-resolution videos (1080P) from HDTF \cite{zhang2021flow} where the pupils are clearly visible. The ground truth 2D pupil centers in these video frames were obtained by using MediaPipe \cite{MediaPipe2019, mediapipe} as the pupil center detector. We then degraded these high-resolution videos to match the resolution (224x224) and quality of VoxCeleb2 \cite{chung2018voxceleb2}, by applying downsampling and blur operations. This process created a realistic test bed for our LightGazeFit, which is designed to process low-resolution video. Both Gaze360 and LightGazeFit were used to process the test videos, and the resulting 3D pupil centers are projected to the image plane to measure $L_2$ error and gaze velocity. The $L_2$ error is defined as the Euclidean distance between the predicted 2D pupil centers and the ground truth, while the gaze velocity is defined as the speed of eyeball movement between adjacent frames, calculated as the 2D movement distance divided by the inter-frame interval of 40 ms.

\begin{figure}
    \centering
    \includegraphics[width=0.48\textwidth]{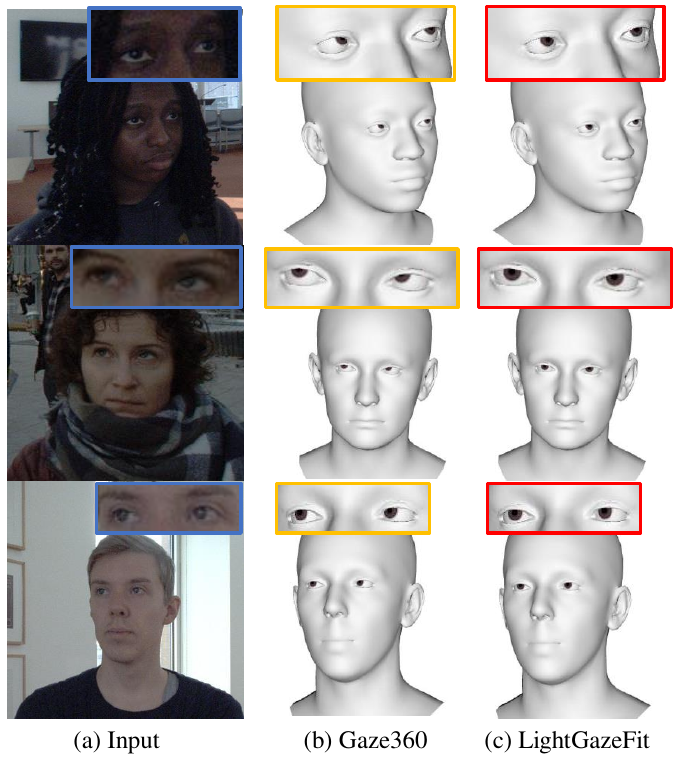}
    \caption{Visual comparisons of 3D eye gaze motions estimated from input facial images by using Gaze360 \cite{kellnhofer2019gaze360} and our method.}
    \label{fig:gaze_compare}
\end{figure}

\begin{table}
\centering
\fontsize{9}{11}\selectfont
\begin{tabular}{c|cc}
	  \hline
	Method & $L_2$ erorr↓  & Velocity \cr
        \hline
        Ground Truth & 0.0000 & 16.2188 \cr
	  \hline
    Gaze360 & 1.4041 & 20.9604 \cr
	LightGazeFit (ours) & \textbf{0.7311} & \textbf{18.5380} \cr
	  \hline
\end{tabular}
\caption{Quantitative comparison of Gaze360 and our LightGazeFit.}
\label{tab:quantitative_evaluation_gaze}
\end{table}

As clearly shown in Tab. \ref{tab:quantitative_evaluation_gaze}, our LightGazeFit achieves superior performance, yielding both a lower prediction error and a gaze velocity that more closely matches the ground truth. 
Fig. \ref{fig:gaze_compare} shows the visual comparison. There exists a deviation from the actual angle of eyeball rotation in Gaze360, indicating that the gaze direction produced by Gaze360 can't be directly converted into the eye pose parameters in FLAME.

\section{Conclusion and Limitations}
In this paper, we study the speech-driven animation of an important facial component, eye gaze, which has been overlooked by the recent research. Through constructing the audio-gaze 3D dataset TKED and the explicit modeling of the translation from speech to 3D eye gaze motion, our proposed TalkingEyes can synthesize eye gaze motion, eye blinks, head motion and facial motion collectively
from speech. Thanks to VAE and VQVAE in learning two separate latent spaces, TalkingEyes can generate pluralistic and natural eye gaze motions and head motions.

Our method has several limitations that we plan to address in future work. 1) The exclusive reliance on speech input prevents our method from reacting to visual elements in a scene, such as interlocutors \cite{S3TOG2024, migCanalesJJ23} or salient objects \cite{SaliencyGazeTVCG2024}. This capability is crucial for fully interactive applications. A promising direction for our future work is to integrate visual inputs, such as the 3D locations of objects and interlocutors, to enable both goal-directed and speech-driven gaze fixation and enhance the model's applicability in interactive virtual environments.
2) Additionally, our method can't produce saccadic eye movements due to the 25 fps framerate of the training dataset. The resulting interval of 40 ms is too coarse to capture the fast dynamics of saccades. Addressing this limitation would require a dataset with a higher temporal resolution, which is a priority for future work. 
3) Different individuals may exhibit different eye and head motion patterns for identical audio input due to factors like personal habit or cultural background. While our method employs a diverse generation strategy to address this variability, it does not support personalized generation for individual users. In future work, we will add speaker identity labels to TKED and use them as conditions during training to learn user-specific styles.
4) Finally, to further improve the physical realism of generated eye gaze in TKED, we will pursue a 3D eye gaze fitting method that incorporates biological constraints \cite{misslisch1998neural} and $L_0$ gradient constraints
\cite{L02014, L02016, PointsL0}.


\bibliographystyle{IEEEtran}
\bibliography{main}

@article{FLAME2017,
  author    = {Tianye Li and
               Timo Bolkart and
               Michael J. Black and
               Hao Li and
               Javier Romero},
  title     = {Learning a model of facial shape and expression from 4D scans},
  journal   = {{ACM} Trans. Graph.},
  volume    = {36},
  number    = {6},
  pages     = {194:1--194:17},
  year      = {2017},
}

@inproceedings{vae2014,
  author       = {Diederik P. Kingma and
                  Max Welling},
  title        = {Auto-Encoding Variational Bayes},
  booktitle    = {Proc. of ICLR},
  year         = {2014}
}

@inproceedings{vae2016,
  author       = {Yunchen Pu and
                  Zhe Gan and
                  Ricardo Henao and
                  Xin Yuan and
                  Chunyuan Li and
                  Andrew Stevens and
                  Lawrence Carin},
  title        = {Variational Autoencoder for Deep Learning of Images, Labels and Captions},
  booktitle    = {Proc. of NIPS},
  pages        = {2352--2360},
  year         = {2016},
}

@inproceedings{GANs2014,
  author       = {Ian J. Goodfellow and
                  Jean Pouget{-}Abadie and
                  Mehdi Mirza and
                  Bing Xu and
                  David Warde{-}Farley and
                  Sherjil Ozair and
                  Aaron C. Courville and
                  Yoshua Bengio},
  title        = {Generative Adversarial Nets},
  booktitle    = {Proc. of NIPS},
  pages        = {2672--2680},
  year         = {2014},
}

@inproceedings{StyleGAN2019,
  author       = {Tero Karras and
                  Samuli Laine and
                  Timo Aila},
  title        = {A Style-Based Generator Architecture for Generative Adversarial Networks},
  booktitle    = {Proc. of CVPR},
  pages        = {4401--4410},
  year         = {2019},
}

@inproceedings{Attention2017,
  author       = {Ashish Vaswani and
                  Noam Shazeer and
                  Niki Parmar and
                  Jakob Uszkoreit and
                  Llion Jones and
                  Aidan N. Gomez and
                  Lukasz Kaiser and
                  Illia Polosukhin},
  title        = {Attention is All you Need},
  booktitle    = {Proc. of NIPS},
  pages        = {5998--6008},
  year         = {2017},
}

@inproceedings{DDPM2020,
  author       = {Jonathan Ho and
                  Ajay Jain and
                  Pieter Abbeel},
  title        = {Denoising Diffusion Probabilistic Models},
  booktitle    = {Proc. of NIPS},
  year         = {2020},
}

@inproceedings{StableDiffusion,
  author       = {Robin Rombach and
                  Andreas Blattmann and
                  Dominik Lorenz and
                  Patrick Esser and
                  Bj{\"{o}}rn Ommer},
  title        = {High-Resolution Image Synthesis with Latent Diffusion Models},
  booktitle    = {Proc. of CVPR},
  pages        = {10674--10685},
  year         = {2022},
}

@inproceedings{FaceDiffuser2023,
  author       = {Stefan Stan and
                  Kazi Injamamul Haque and
                  Zerrin Yumak},
  title        = {FaceDiffuser: Speech-Driven 3D Facial Animation Synthesis Using Diffusion},
  booktitle    = {{ACM} Conference on Motion, Interaction and Games},
  pages        = {13:1--13:11},
  year         = {2023},
}

@inproceedings{SadTalker2023,
  author       = {Wenxuan Zhang and
                  Xiaodong Cun and
                  Xuan Wang and
                  Yong Zhang and
                  Xi Shen and
                  Yu Guo and
                  Ying Shan and
                  Fei Wang},
  title        = {SadTalker: Learning Realistic 3D Motion Coefficients for Stylized
                  Audio-Driven Single Image Talking Face Animation},
  booktitle    = {Proc. of CVPR},
  pages        = {8652--8661},
  year         = {2023},
}

@inproceedings{VOCA2019,
  author       = {Daniel Cudeiro and
                  Timo Bolkart and
                  Cassidy Laidlaw and
                  Anurag Ranjan and
                  Michael J. Black},
  title        = {Capture, Learning, and Synthesis of 3D Speaking Styles},
  booktitle    = {Proc. of CVPR},
  pages        = {10101--10111},
  year         = {2019},
}

@inproceedings{MeshTalk2021,
  author       = {Alexander Richard and
                  Michael Zollh{\"{o}}fer and
                  Yandong Wen and
                  Fernando De la Torre and
                  Yaser Sheikh},
  title        = {MeshTalk: 3D Face Animation from Speech using Cross-Modality Disentanglement},
  booktitle    = {Proc. of ICCV},
  pages        = {1153--1162},
  year         = {2021},
}

@inproceedings{FaceFormer2022,
  author       = {Yingruo Fan and
                  Zhaojiang Lin and
                  Jun Saito and
                  Wenping Wang and
                  Taku Komura},
  title        = {FaceFormer: Speech-Driven 3D Facial Animation with Transformers},
  booktitle    = {Proc. of CVPR},
  pages        = {18749--18758},
  year         = {2022},
}

@inproceedings{CodeTalker2023,
  author       = {Jinbo Xing and
                  Menghan Xia and
                  Yuechen Zhang and
                  Xiaodong Cun and
                  Jue Wang and
                  Tien{-}Tsin Wong},
  title        = {CodeTalker: Speech-Driven 3D Facial Animation with Discrete Motion
                  Prior},
  booktitle    = {Proc. of CVPR},
  pages        = {12780--12790},
  year         = {2023},
}

@inproceedings{VQVAE2017,
  author       = {A{\"{a}}ron van den Oord and
                  Oriol Vinyals and
                  Koray Kavukcuoglu},
  title        = {Neural Discrete Representation Learning},
  booktitle    = {Proc. of NIPS},
  pages        = {6306--6315},
  year         = {2017}
}

@inproceedings{wav2vec2020,
  author       = {Alexei Baevski and
                  Yuhao Zhou and
                  Abdelrahman Mohamed and
                  Michael Auli},
  title        = {wav2vec 2.0: {A} Framework for Self-Supervised Learning of Speech
                  Representations},
  booktitle    = {Proc. of NIPS},
  year         = {2020},
}

@inproceedings{Deep3DFace2019,
  author       = {Yu Deng and
                  Jiaolong Yang and
                  Sicheng Xu and
                  Dong Chen and
                  Yunde Jia and
                  Xin Tong},
  title        = {Accurate 3D Face Reconstruction With Weakly-Supervised Learning: From
                  Single Image to Image Set},
  booktitle    = {Proc. of CVPR Workshops},
  pages        = {285--295},
  year         = {2019},
}

@article{DECA2021,
  author       = {Yao Feng and
                  Haiwen Feng and
                  Michael J. Black and
                  Timo Bolkart},
  title        = {Learning an animatable detailed 3D face model from in-the-wild images},
  journal      = {{ACM} Trans. Graph.},
  volume       = {40},
  number       = {4},
  pages        = {88:1--88:13},
  year         = {2021},
}

@inproceedings{EMOCA2022,
  author       = {Radek Danecek and
                  Michael J. Black and
                  Timo Bolkart},
  title        = {{EMOCA:} Emotion Driven Monocular Face Capture and Animation},
  booktitle    = {Proc. of CVPR},
  pages        = {20279--20290},
  year         = {2022},
}

@article{LipreadVideo2022,
  author       = {Panagiotis Paraskevas Filntisis and
                  George Retsinas and
                  Foivos Paraperas Papantoniou and
                  Athanasios Katsamanis and
                  Anastasios Roussos and
                  Petros Maragos},
  title        = {Visual Speech-Aware Perceptual 3D Facial Expression Reconstruction from Videos},
  journal      = {arXiv preprint arXiv:2207.11094},
  year         = {2022},
}

@article{BIWI2010,
  author       = {Gabriele Fanelli and
                  J{\"{u}}rgen Gall and
                  Harald Romsdorfer and
                  Thibaut Weise and
                  Luc Van Gool},
  title        = {A 3-D Audio-Visual Corpus of Affective Communication},
  journal      = {{IEEE} Trans. Multim.},
  volume       = {12},
  number       = {6},
  pages        = {591--598},
  year         = {2010},
}

@article{chung2018voxceleb2,
  title={Voxceleb2: Deep speaker recognition},
  author={Chung, Joon Son and Nagrani, Arsha and Zisserman, Andrew},
  journal={arXiv preprint arXiv:1806.05622},
  year={2018}
}

@inproceedings{zhang2021flow,
  title={Flow-guided one-shot talking face generation with a high-resolution audio-visual dataset},
  author={Zhang, Zhimeng and Li, Lincheng and Ding, Yu and Fan, Changjie},
  booktitle={Proc. of CVPR},
  pages={3661--3670},
  year={2021}
}

@article{wang2019realtime,
  author       = {Zhiyong Wang and
                  Jinxiang Chai and
                  Shihong Xia},
  title        = {Realtime and Accurate 3D Eye Gaze Capture with DCNN-Based Iris and Pupil Segmentation},
  journal      = {{IEEE} Trans. Vis. Comput. Graph.},
  volume       = {27},
  number       = {1},
  pages        = {190--203},
  year         = {2021}
}

@inproceedings{he2023speech4mesh,
  title={Speech4mesh: Speech-assisted monocular 3d facial reconstruction for speech-driven 3d facial animation},
  author={He, Shan and He, Haonan and Yang, Shuo and Wu, Xiaoyan and Xia, Pengcheng and Yin, Bing and Liu, Cong and Dai, Lirong and Xu, Chang},
  booktitle={Proc. of CVPR},
  pages={14192--14202},
  year={2023}
}

@article{DiffPoseTalkTOG2024,
  author       = {Zhiyao Sun and
                  Tian Lv and
                  Sheng Ye and
                  Matthieu Gaetan Lin and
                  Jenny Sheng and
                  Yu{-}Hui Wen and
                  Minjing Yu and
                  Yong{-}Jin Liu},
  title        = {DiffPoseTalk: Speech-Driven Stylistic 3D Facial Animation and Head
                  Pose Generation via Diffusion Models},
  journal      = {{ACM} Trans. Graph.},
  volume       = {43},
  number       = {4},
  pages        = {46:1--46:9},
  year         = {2024}
}

@inproceedings{yang2024probabilistic,
  title={Probabilistic Speech-Driven 3D Facial Motion Synthesis: New Benchmarks Methods and Applications},
  author={Yang, Karren D and Ranjan, Anurag and Chang, Jen-Hao Rick and Vemulapalli, Raviteja and Tuzel, Oncel},
  booktitle={Proc. of CVPR},
  pages={27294--27303},
  year={2024}
}

@inproceedings{wang2020mead,
  title={Mead: A large-scale audio-visual dataset for emotional talking-face generation},
  author={Wang, Kaisiyuan and Wu, Qianyi and Song, Linsen and Yang, Zhuoqian and Wu, Wayne and Qian, Chen and He, Ran and Qiao, Yu and Loy, Chen Change},
  booktitle={Proc. of ECCV},
  pages={700--717},
  year={2020}
}

@inproceedings{zielonka2022towards,
  title={Towards metrical reconstruction of human faces},
  author={Zielonka, Wojciech and Bolkart, Timo and Thies, Justus},
  booktitle={Proc. of ECCV},
  pages={250--269},
  year={2022},
}

@inproceedings{kellnhofer2019gaze360,
  title={Gaze360: Physically unconstrained gaze estimation in the wild},
  author={Kellnhofer, Petr and Recasens, Adria and Stent, Simon and Matusik, Wojciech and Torralba, Antonio},
  booktitle={Proc. of ICCV},
  pages={6912--6921},
  year={2019}
}

@article{wang2016realtime,
  title        = {Realtime 3D eye gaze animation using a single {RGB} camera},
  author       = {Congyi Wang and
                  Fuhao Shi and
                  Shihong Xia and
                  Jinxiang Chai},
  journal      = {{ACM} Trans. Graph.},
  volume       = {35},
  number       = {4},
  pages        = {118:1--118:14},
  year         = {2016},
}

@article{wen2017real,
  author       = {Quan Wen and
                  Feng Xu and
                  Jun{-}Hai Yong},
  title        = {Real-Time 3D Eye Performance Reconstruction for {RGBD} Cameras},
  journal      = {{IEEE} Trans. Vis. Comput. Graph.},
  volume       = {23},
  number       = {12},
  pages        = {2586--2598},
  year         = {2017}
}

@inproceedings{zhang2015appearance,
  title={Appearance-based gaze estimation in the wild},
  author={Zhang, Xucong and Sugano, Yusuke and Fritz, Mario and Bulling, Andreas},
  booktitle={Proc. of CVPR},
  pages={4511--4520},
  year={2015}
}

@article{cheng2020gaze,
  author       = {Yihua Cheng and
                  Xucong Zhang and
                  Feng Lu and
                  Yoichi Sato},
  title        = {Gaze Estimation by Exploring Two-Eye Asymmetry},
  journal      = {{IEEE} Trans. Image Process.},
  volume       = {29},
  pages        = {5259--5272},
  year         = {2020},
}

@inproceedings{park2018deep,
  title={Deep pictorial gaze estimation},
  author={Park, Seonwook and Spurr, Adrian and Hilliges, Otmar},
  booktitle={Proc. of ECCV},
  pages={721--738},
  year={2018}
}

@inproceedings{park2019few,
  title={Few-shot adaptive gaze estimation},
  author={Park, Seonwook and Mello, Shalini De and Molchanov, Pavlo and Iqbal, Umar and Hilliges, Otmar and Kautz, Jan},
  booktitle={Proceedings of the IEEE/CVF international conference on computer vision},
  pages={9368--9377},
  year={2019}
}

@inproceedings{yu2019improving,
  title={Improving few-shot user-specific gaze adaptation via gaze redirection synthesis},
  author={Yu, Yu and Liu, Gang and Odobez, Jean-Marc},
  booktitle={Proc. of CVPR},
  pages={11937--11946},
  year={2019}
}

@article{S3TOG2024,
  author       = {Yifang Pan and
                  Rishabh Agrawal and
                  Karan Singh},
  title        = {{S3:} Speech, Script and Scene driven Head and Eye Animation},
  journal      = {{ACM} Trans. Graph.},
  volume       = {43},
  number       = {4},
  pages        = {47:1--47:12},
  year         = {2024}
}

@article{EyeMotion2019,
  author       = {Aobo Jin and
                  Qixin Deng and
                  Yuting Zhang and
                  Zhigang Deng},
  title        = {A Deep Learning-Based Model for Head and Eye Motion Generation in Three-party Conversations},
  journal      = {Proc. {ACM} Comput. Graph. Interact. Tech.},
  volume       = {2},
  number       = {2},
  pages        = {9:1--9:19},
  year         = {2019},
}

@article{SaliencyGazeTVCG2024,
  author       = {Ific Goud{\'{e}} and
                  Alexandre Bruckert and
                  Anne{-}H{\'{e}}l{\`{e}}ne Olivier and
                  Julien Pettr{\'{e}} and
                  R{\'{e}}mi Cozot and
                  Kadi Bouatouch and
                  Marc Christie and
                  Ludovic Hoyet},
  title        = {Real-Time Multi-Map Saliency-Driven Gaze Behavior for Non-Conversational Characters},
  journal      = {{IEEE} Trans. Vis. Comput. Graph.},
  volume       = {30},
  number       = {7},
  pages        = {3871--3883},
  year         = {2024},
}

@article{3DTalkingFaceHead2023,
  author       = {Chenxu Zhang and
                  Saifeng Ni and
                  Zhipeng Fan and
                  Hongbo Li and
                  Ming Zeng and
                  Madhukar Budagavi and
                  Xiaohu Guo},
  title        = {3D Talking Face With Personalized Pose Dynamics},
  journal      = {{IEEE} Trans. Vis. Comput. Graph.},
  volume       = {29},
  number       = {2},
  pages        = {1438--1449},
  year         = {2023}
}

@inproceedings{EMOTE2023,
  author       = {Radek Danecek and
                  Kiran Chhatre and
                  Shashank Tripathi and
                  Yandong Wen and
                  Michael J. Black and
                  Timo Bolkart},
  title        = {Emotional Speech-Driven Animation with Content-Emotion Disentanglement},
  booktitle    = {Proc. of SIGGRAPH Asia},
  pages        = {41:1--41:13},
  year         = {2023},
}

@inproceedings{EmoTalk2023,
  author       = {Ziqiao Peng and
                  Haoyu Wu and
                  Zhenbo Song and
                  Hao Xu and
                  Xiangyu Zhu and
                  Jun He and
                  Hongyan Liu and
                  Zhaoxin Fan},
  title        = {EmoTalk: Speech-Driven Emotional Disentanglement for 3D Face Animation},
  booktitle    = {Proc. of ICCV},
  pages        = {20630--20640},
  year         = {2023},
}

@inproceedings{Media2FaceSIG2024,
  author       = {Qingcheng Zhao and
                  Pengyu Long and
                  Qixuan Zhang and
                  Dafei Qin and
                  Han Liang and
                  Longwen Zhang and
                  Yingliang Zhang and
                  Jingyi Yu and
                  Lan Xu},
  title        = {Media2Face: Co-speech Facial Animation Generation With Multi-Modality Guidance},
  booktitle    = {Proc. of SIGGRAPH},
  pages        = {18},
  year         = {2024}
}

@article{HeadEyeAnimation2007,
  author       = {Soh Masuko and
                  Junichi Hoshino},
  title        = {Head-eye Animation Corresponding to a Conversation for {CG} Characters},
  journal      = {Comput. Graph. Forum},
  volume       = {26},
  number       = {3},
  pages        = {303--312},
  year         = {2007}
}

@inproceedings{VirtualCharacterSCA2013,
  author       = {Stacy Marsella and
                  Yuyu Xu and
                  Margaux Lhommet and
                  Andrew W. Feng and
                  Stefan Scherer and
                  Ari Shapiro},
  title        = {Virtual character performance from speech},
  booktitle    = {Proc. of SCA},
  pages        = {25--35},
  year         = {2013}
}

@article{ConversationalAgents2012,
  author       = {Soroosh Mariooryad and
                  Carlos Busso},
  title        = {Generating Human-Like Behaviors Using Joint, Speech-Driven Models for Conversational Agents},
  journal      = {{IEEE} Trans. Speech Audio Process.},
  volume       = {20},
  number       = {8},
  pages        = {2329--2340},
  year         = {2012}
}

@article{LiveSpeehEyeTVCG2012,
  author       = {Binh Huy Le and
                  Xiaohan Ma and
                  Zhigang Deng},
  title        = {Live Speech Driven Head-and-Eye Motion Generators},
  journal      = {{IEEE} Trans. Vis. Comput. Graph.},
  volume       = {18},
  number       = {11},
  pages        = {1902--1914},
  year         = {2012}
}

@inproceedings{liu2024emoface,
  title={EmoFace: Audio-driven Emotional 3D Face Animation},
  author={Liu, Chang and Lin, Qunfen and Zeng, Zijiao and Pan, Ye},
  booktitle={Proc. of  Virtual Reality},
  pages={387--397},
  year={2024},
}

@article{cech2016real,
  title={Real-time eye blink detection using facial landmarks},
  author={Tereza Soukupova and Jan Cech},
  journal={Cent. Mach. Perception, Dep. Cybern. Fac. Electr. Eng. Czech Tech. Univ. Prague},
  pages={1--8},
  year={2016}
}

@article{EyeGazeReview2015,
  author       = {Kerstin Ruhland and
                  Christopher E. Peters and
                  Sean Andrist and
                  Jeremy B. Badler and
                  Norman I. Badler and
                  Michael Gleicher and
                  Bilge Mutlu and
                  Rachel McDonnell},
  title        = {A Review of Eye Gaze in Virtual Agents, Social Robotics and {HCI:}
                  Behaviour Generation, User Interaction and Perception},
  journal      = {Comput. Graph. Forum},
  volume       = {34},
  number       = {6},
  pages        = {299--326},
  year         = {2015}
}

@article{GazeEstimationSurvey2024,
  author       = {Yihua Cheng and
                  Haofei Wang and
                  Yiwei Bao and
                  Feng Lu},
  title        = {Appearance-Based Gaze Estimation With Deep Learning: {A} Review and Benchmark},
  journal      = {{IEEE} Trans. Pattern Anal. Mach. Intell.},
  volume       = {46},
  number       = {12},
  pages        = {7509--7528},
  year         = {2024}
}

@article{JALI2016,
  author       = {Pif Edwards and
                  Chris Landreth and
                  Eugene Fiume and
                  Karan Singh},
  title        = {{JALI:} an animator-centric viseme model for expressive lip synchronization},
  journal      = {{ACM} Trans. Graph.},
  volume       = {35},
  number       = {4},
  pages        = {127:1--127:11},
  year         = {2016}
}

@inproceedings{AnimatedSpeech2001,
  author       = {Michael M. Cohen and
                  Rashid Clark and
                  Dominic W. Massaro},
  title        = {Animated speech: research progress and applications},
  booktitle    = {Proc. of Auditory-Visual Speech Processing},
  pages        = {200},
  year         = {2001},
}

@inproceedings{DynamicUnits2012,
  author       = {Sarah L. Taylor and
                  Moshe Mahler and
                  Barry{-}John Theobald and
                  Iain A. Matthews},
  title        = {Dynamic Units of Visual Speech},
  booktitle    = {Proc. of SCA},
  pages        = {275--284},
  year         = {2012},
}

@inproceedings{LipSyncGames2013,
  author       = {Yuyu Xu and
                  Andrew W. Feng and
                  Stacy Marsella and
                  Ari Shapiro},
  title        = {A Practical and Configurable Lip Sync Method for Games},
  booktitle    = {{ACM} Conference on Motion, Interaction and Games},
  pages        = {131--140},
  year         = {2013},
}

@article{Ophthal1972,
  author       = {E Bizzi and R E Kalil and P Morasso and V Tagliasco},
  title        = {Central programming and peripheral feedback
during eye-head coordination in monkeys},
  journal      = {Bibl. Ophthal.},
  volume       = {82},
  pages        = {220--232},
  year         = {1972}
}

@article{Warabi1977,
  author       = {Tateo Warabi},
  title        = {The reaction time of eye-head coordination in man},
  journal      = {Neurosci. Lett.},
  volume       = {6},
  pages        = {47--51},
  year         = {1977}
}

@article{GestureDiffuCLIP2023,
  author       = {Tenglong Ao and
                  Zeyi Zhang and
                  Libin Liu},
  title        = {GestureDiffuCLIP: Gesture Diffusion Model with {CLIP} Latents},
  journal      = {{ACM} Trans. Graph.},
  volume       = {42},
  number       = {4},
  pages        = {42:1--42:18},
  year         = {2023}
}

@inproceedings{EMAGE2024,
  author       = {Haiyang Liu and
                  Zihao Zhu and
                  Giorgio Becherini and
                  Yichen Peng and
                  Mingyang Su and
                  You Zhou and
                  Xuefei Zhe and
                  Naoya Iwamoto and
                  Bo Zheng and
                  Michael J. Black},
  title        = {{EMAGE:} Towards Unified Holistic Co-Speech Gesture Generation via Expressive Masked Audio Gesture Modeling},
  booktitle    = {Proc. of CVPR},
  pages        = {1144--1154},
  year         = {2024}
}

@inproceedings{TalkShow2023,
  author       = {Hongwei Yi and
                  Hualin Liang and
                  Yifei Liu and
                  Qiong Cao and
                  Yandong Wen and
                  Timo Bolkart and
                  Dacheng Tao and
                  Michael J. Black},
  title        = {Generating Holistic 3D Human Motion from Speech},
  booktitle    = {Proc. of CVPR},
  pages        = {469--480},
  year         = {2023}
}

@inproceedings{DiffSFSR2024,
  author       = {Renshuai Liu and
                  Bowen Ma and
                  Wei Zhang and
                  Zhipeng Hu and
                  Changjie Fan and
                  Tangjie Lv and
                  Yu Ding and
                  Xuan Cheng},
  title        = {Towards a Simultaneous and Granular Identity-Expression Control in
                  Personalized Face Generation},
  booktitle    = {Proc. of CVPR},
  pages        = {2114--2123},
  year         = {2024}
}

@inproceedings{DNPMICME2024,
  author       = {Haitao Cao and
                  Baoping Cheng and
                  Qiran Pu and
                  Haocheng Zhang and
                  Bin Luo and
                  Yixiang Zhuang and
                  Juncong Lin and
                  Liyan Chen and
                  Xuan Cheng},
  title        = {{DNPM:} {A} Neural Parametric Model for the Synthesis of Facial Geometric Details},
  booktitle    = {Proc. of ICME},
  pages        = {1--6},
  year         = {2024},
}

@inproceedings{VQVAE2-2019,
  author       = {Ali Razavi and
                  A{\"{a}}ron van den Oord and
                  Oriol Vinyals},
  title        = {Generating Diverse High-Fidelity Images with {VQ-VAE-2}},
  booktitle    = {Proc. of NIPS},
  pages        = {14837--14847},
  year         = {2019}
}

@article{3DGSTOG2023,
  author       = {Bernhard Kerbl and
                  Georgios Kopanas and
                  Thomas Leimk{\"{u}}hler and
                  George Drettakis},
  title        = {3D Gaussian Splatting for Real-Time Radiance Field Rendering},
  journal      = {{ACM} Trans. Graph.},
  volume       = {42},
  number       = {4},
  pages        = {139:1--139:14},
  year         = {2023}
}

@inproceedings{GaussianAvatarsCVPR2024,
  author       = {Shenhan Qian and
                  Tobias Kirschstein and
                  Liam Schoneveld and
                  Davide Davoli and
                  Simon Giebenhain and
                  Matthias Nie{\ss}ner},
  title        = {GaussianAvatars: Photorealistic Head Avatars with Rigged 3D Gaussians},
  booktitle    = {Proc. of CVPR},
  pages        = {20299--20309},
  year         = {2024}
}

@article{ICTPAMI2024,
  author       = {Ziyu Wan and
                  Jingbo Zhang and
                  Dongdong Chen and
                  Jing Liao},
  title        = {High-Fidelity and Efficient Pluralistic Image Completion With Transformers},
  journal      = {{IEEE} Trans. Pattern Anal. Mach. Intell.},
  volume       = {46},
  number       = {12},
  pages        = {9612--9629},
  year         = {2024}
}

@inproceedings{PICCVPR2019,
  author       = {Chuanxia Zheng and
                  Tat{-}Jen Cham and
                  Jianfei Cai},
  title        = {Pluralistic Image Completion},
  booktitle    = {Proc. of CVPR},
  pages        = {1438--1447},
  year         = {2019},
}

@article{MediaPipe2019,
  author       = {Camillo Lugaresi and
                  Jiuqiang Tang and
                  Hadon Nash and
                  Chris McClanahan and
                  Esha Uboweja and
                  Michael Hays and
                  Fan Zhang and
                  Chuo{-}Ling Chang and
                  Ming Guang Yong and
                  Juhyun Lee and
                  Wan{-}Teh Chang and
                  Wei Hua and
                  Manfred Georg and
                  Matthias Grundmann},
  title        = {MediaPipe: {A} Framework for Building Perception Pipelines},
  journal      = {arXiv preprint arXiv:1906.08172},
  year         = {2019},
}

@misc{mediapipe,
 author = {Google},
 title = {MediaPipe},
howpublished={\url{https://github.com/google-ai-edge/mediapipe}},
}

@inproceedings{migCanalesJJ23,
  author       = {Ryan Canales and
                  Eakta Jain and
                  Sophie J{\"{o}}rg},
  title        = {Real-Time Conversational Gaze Synthesis for Avatars},
  booktitle    = {{ACM} Conference on Motion, Interaction and Games},
  pages        = {17:1--17:7},
  year         = {2023},
}

@article{misslisch1998neural,
  title={Neural constraints on eye motion in human eye-head saccades},
  author={Misslisch, H and Tweed, D and Vilis, T},
  journal={Journal of Neurophysiology},
  volume={79},
  number={2},
  pages={859--869},
  year={1998},
}

@article{FaceRefiner2024,
  author       = {Chengyang Li and
                  Baoping Cheng and
                  Yao Cheng and
                  Haocheng Zhang and
                  Renshuai Liu and
                  Yinglin Zheng and
                  Jing Liao and
                  Xuan Cheng},
  title        = {FaceRefiner: High-Fidelity Facial Texture Refinement With Differentiable
                  Rendering-Based Style Transfer},
  journal      = {{IEEE} Trans. Multim.},
  volume       = {26},
  pages        = {7225--7236},
  year         = {2024}
}

@article{DCT2024,
  author       = {Renshuai Liu and
                  Yao Cheng and
                  Sifei Huang and
                  Chengyang Li and
                  Xuan Cheng},
  title        = {Transformer-Based High-Fidelity Facial Displacement Completion for
                  Detailed 3D Face Reconstruction},
  journal      = {{IEEE} Trans. Multim.},
  volume       = {26},
  pages        = {799--810},
  year         = {2024}
}

@article{PointsL0,
  author       = {Xuan Cheng and
                  Ming Zeng and
                  Jinpeng Lin and
                  Zizhao Wu and
                  Xinguo Liu},
  title        = {Efficient L0 resampling of point sets},
  journal      = {Comput. Aided Geom. Des.},
  volume       = {75},
  year         = {2019},

}

@article{Learn2Talk2025,
  author       = {Yixiang Zhuang and
                  Baoping Cheng and
                  Yao Cheng and
                  Yuntao Jin and
                  Renshuai Liu and
                  Chengyang Li and
                  Xuan Cheng and
                  Jing Liao and
                  Juncong Lin},
  title        = {Learn2Talk: 3D Talking Face Learns From 2D Talking Face},
  journal      = {{IEEE} Trans. Vis. Comput. Graph.},
  volume       = {31},
  number       = {9},
  pages        = {5829--5841},
  year         = {2025}
}

@article{L02014,
  author       = {Xuan Cheng and
                  Ming Zeng and
                  Xinguo Liu},
  title        = {Feature-preserving filtering with L0 gradient minimization},
  journal      = {Comput. Graph.},
  volume       = {38},
  pages        = {150--157},
  year         = {2014},
}

@article{L02016,
  author       = {Xuan Cheng and
                  Yuanli Feng and
                  Ming Zeng and
                  Xinguo Liu},
  title        = {Video segmentation with L0 gradient minimization},
  journal      = {Comput. Graph.},
  volume       = {54},
  pages        = {38--46},
  year         = {2016},
}

\end{document}